\documentclass[10pt,twocolumn,letterpaper]{article}

\usepackage{cvpr}      

\usepackage{lineno}

\usepackage{pgffor}      
\usepackage{adjustbox}   
\usepackage{ragged2e}    
\usepackage{caption}     
\usepackage{afterpage}
\usepackage{cuted} 
\usepackage{graphicx}

\usepackage{algorithm2e}  
\usepackage{amsmath}
\usepackage{amssymb}

\immediate\write18{ls images/*.jpg images/*.png images/*.jpeg > imagelist.txt}

\definecolor{cvprblue}{rgb}{0.21,0.49,0.74}
\usepackage[pagebackref,breaklinks,colorlinks,allcolors=cvprblue]{hyperref}

\title{Towards Open-Vocabulary Industrial Defect Understanding with a Large-Scale Multimodal Dataset}

\author{TsaiChing Ni \quad ZhenQi Chen \quad YuanFu Yang\\
Institute of Intelligent Systems, National Yang Ming Chiao Tung University
}

\begin{document}
\maketitle

\maketitle

\begin{strip}
    \centering
    \includegraphics[width=1\textwidth]{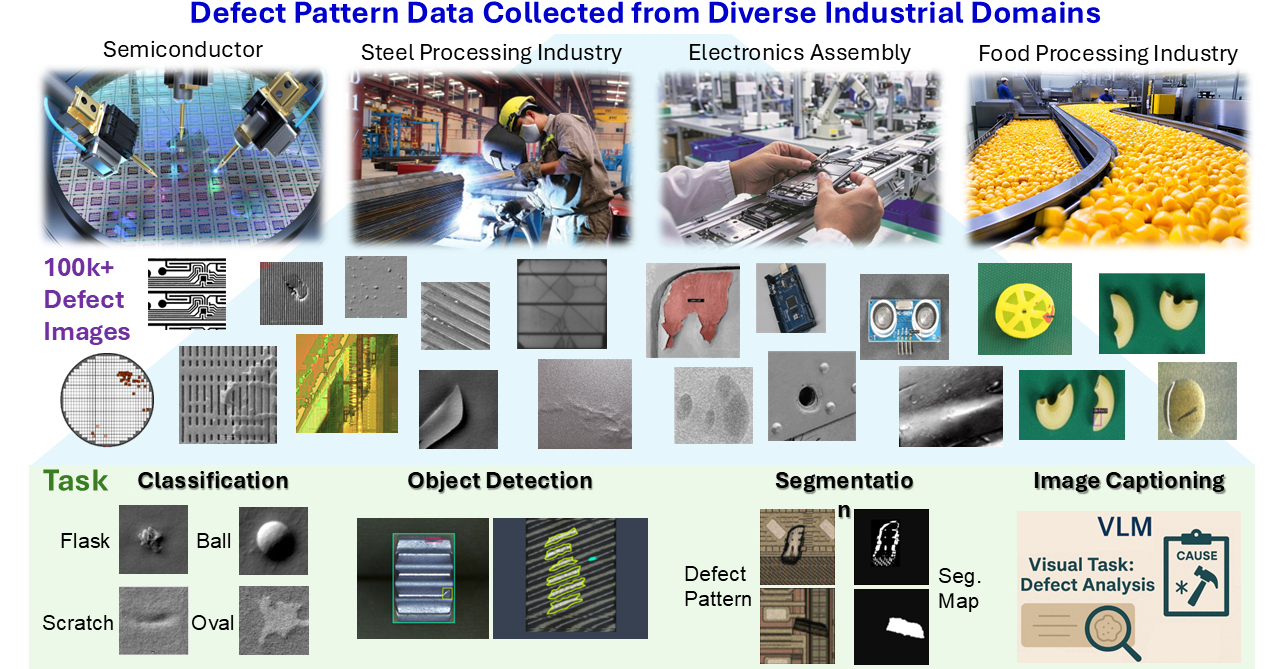} 
    \vspace{-0.8em}
    \captionof{figure}{Overview of the proposed IMDD-1M dataset, its diverse industrial domains, corresponding downstream tasks, and potential extensions to vision-language model applications.}
    \label{fig:teaser}
    \vspace{1em}
\end{strip}

\begin{abstract}
We present \textbf{IMDD-1M}, the first large-scale Industrial Multimodal Defect Dataset comprising 1,000,000 aligned image-text pairs, designed to advance multimodal learning for manufacturing and quality inspection. IMDD-1M contains high-resolution real-world defects spanning over 60 material categories and more than 400 defect types, each accompanied by expert-verified annotations and fine-grained textual descriptions detailing defect location, severity, and contextual attributes. This dataset enables a wide spectrum of applications, including classification, segmentation, retrieval, captioning, and generative modeling. Building upon IMDD-1M, we train a diffusion-based vision-language foundation model from scratch, specifically tailored for industrial scenarios. The model serves as a generalizable foundation that can be efficiently adapted to specialized domains through lightweight fine-tuning. With less than 5\% of the task-specific data required by dedicated expert models, it achieves comparable performance, highlighting the potential of data-efficient foundation model adaptation for industrial inspection and generation, paving the way for scalable, domain-adaptive, and knowledge-grounded manufacturing intelligence.
\looseness=-1
\enlargethispage{1\baselineskip}
\nopagebreak
\end{abstract}    
\section{Introduction}
\label{sec:introduction}

\hspace{1em}Industrial defect detection is critical for ensuring product quality and operational efficiency in modern manufacturing. Traditional manual inspection suffers from high labor costs, subjective judgment, and limited throughput, driving widespread adoption of automated optical inspection (AOI) systems across semiconductor, electronics, and precision manufacturing sectors. Despite significant advances, current AOI systems remain constrained by high false alarm rates, poor adaptability to novel defect patterns, and inability to generalize across diverse manufacturing contexts. To address these limitations, we introduce IMDD-1M, a large-scale industrial defect dataset with multimodal annotations, along with a suite of downstream tasks and VLM applications (as shown in Figure \ref{fig:teaser}).

\begin{figure*}[t]
\centering
\includegraphics[width=\textwidth]{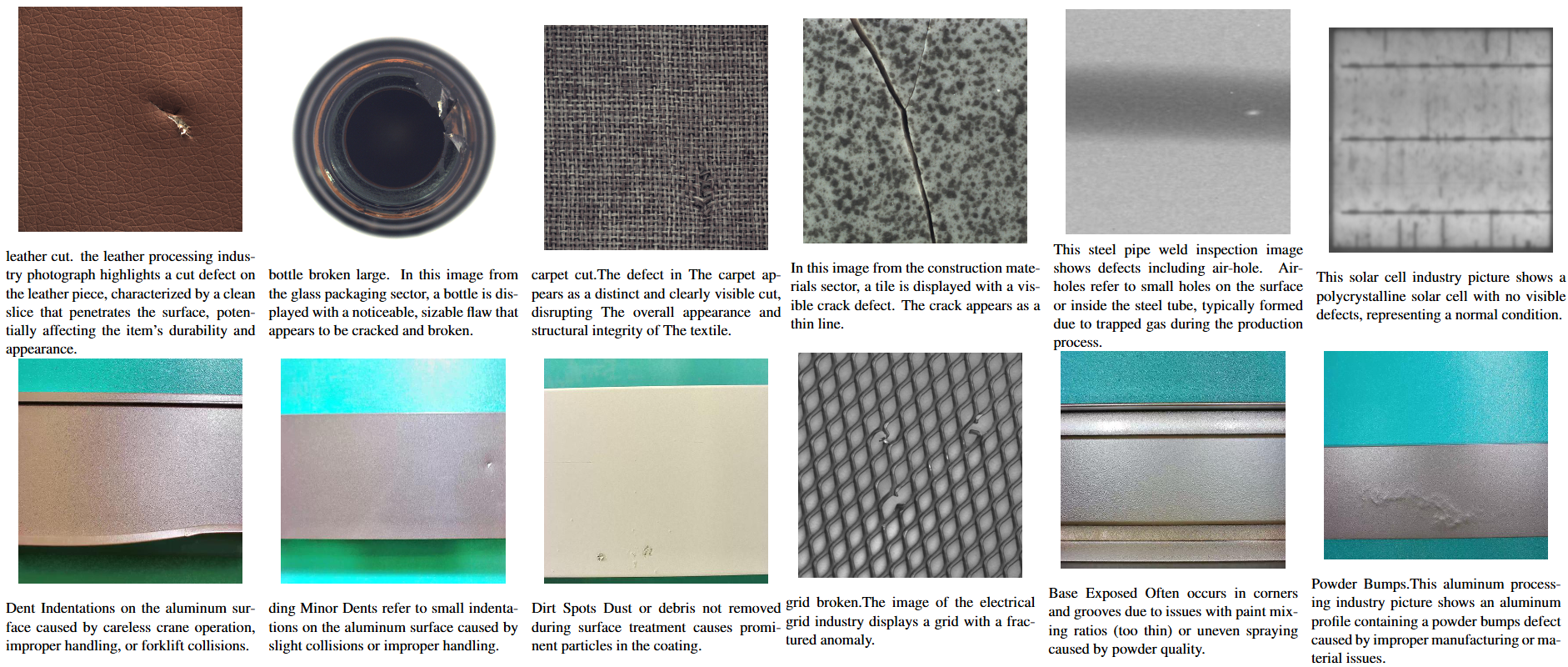}
\vspace{-0.7em}
\caption{Illustrative overview of IMDD-1M showing diverse image-text pairs across multiple industrial domains, each with expert-verified annotations capturing fine-grained defect types, materials, and manufacturing contexts. The dataset serves as a large-scale foundation for vision-language modeling in industrial inspection.}
\label{fig:over}
\end{figure*}
Existing deep learning approaches further expose these limitations. Specialized architectures such as You Only Look Once (YOLO)-based detectors excel at specific tasks but lack unified multi-task capabilities, require extensive pixel-level annotations, struggle with rare defects, and operate as black-box discriminators without semantic interpretability. Recent VLMs such as Contrastive Language-Image Pre-training (CLIP)~\cite{clip}, ALIGN~\cite{align}, and Flamingo~\cite{flamingo} have revolutionized natural image understanding by aligning visual and textual semantics. However, industrial defects are subtle, localized, and require domain-specific terminology (e.g., "delamination," "solder void"). Trained predominantly on natural images, existing VLMs lack specialized knowledge for industrial visual-semantic correlations.

To bridge this gap, we introduce IMDD-1M, the first million-scale industrial defect dataset with aligned image-text pairs, together with a diffusion-based multimodal foundation model that unifies generative and discriminative capabilities for defect generation, segmentation, detection, and semantic grounding. Our key contributions are:

\textbf{(1)} IMDD-1M, a million-scale industrial dataset spanning 421 defect types across 63 manufacturing domains (automotive, electronics, metals, textiles, packaging). Each sample contains expert-verified image-text annotations, surpassing existing benchmarks by approximately two orders of magnitude.

\textbf{(2)} A hybrid annotation pipeline combining expert verification with Large Language Model (LLM)-assisted caption generation, ensuring domain-specific technical accuracy and linguistic consistency across diverse industrial terminologies.

\textbf{(3)} A diffusion-based multimodal foundation model trained on IMDD-1M that integrates discriminative capabilities (segmentation and detection) with generative ones (synthesis and augmentation) within a single architecture, demonstrating effective performance in industrial anomaly understanding.

\textbf{(4)} Comprehensive evaluation protocols and benchmarks for defect detection, segmentation, and synthesis, providing standardized metrics for systematic assessment in practical industrial environments.

\section{Related Work}
\label{sec:formatting}
\hspace{1em}Automated defect detection has evolved through benchmark datasets of increasing scale and complexity. Early efforts like DAGM~\cite{DAGM} and KolektorSDD~\cite{KolektorSDD} provided limited synthetic or single-domain data. MVTec AD~\cite{MVTecAD} introduced pixel-level annotations across multiple categories but reached performance saturation. Subsequent datasets including BTAD~\cite{BTAD}, VisA~\cite{VisA}, and Real-IAD~\cite{RealIAD} progressively improved realism and diversity, while domain-specific datasets emerged for steel~\cite{NEUDET}, X-ray inspection~\cite{DVXRAY}, circuit boards~\cite{PCB}, and other applications~\cite{mvtecad2,magnetictile,crack}. However, all remain constrained by limited scale and lack multimodal annotations.

Natural image datasets such as ImageNet~\cite{ImageNet}, COCO~\cite{COCO}, and LAION~\cite{LAION} enabled advances in VLMs but exhibit fundamental mismatches with industrial needs: defects are subtle and localized, requiring specialized terminology and pixel-level precision. Our IMDD-1M bridges this gap as the first large-scale industrial dataset with image-text pairs, enabling multimodal learning for defect analysis (as shown in Table~\ref{tab:dataset_comp}).

\begin{table}[t]
\centering
\caption{
Comparison of industrial defect datasets.
Unlike previous datasets, IMDD-1M introduces large-scale image-text pairs, enabling multimodal learning in the industrial domain.
}
\label{tab:dataset_comp}
\resizebox{\columnwidth}{!}{
\begin{tabular}{l|c|c|c|c}
\toprule
\textbf{Dataset} & \textbf{Year} & \textbf{\# Images} & \textbf{\# Domains} & \textbf{Text Annotations} \\
\midrule
DAGM~\cite{DAGM} & 2016 & 1.5K & 1 (Synthetic) & No \\
KolektorSDD~\cite{KolektorSDD} & 2019 & 400 & 1 (Electronics) & No \\
MVTec AD~\cite{MVTecAD} & 2019 & 5.4K & 15 (Objects) & No \\
BTAD~\cite{BTAD} & 2021 & 2.5K & 3 (Mixed) & No \\
VisA~\cite{VisA} & 2022 & 10.8K & 12 (Packaging) & No \\
Real-IAD~\cite{RealIAD} & 2024 & 67K & 30 (Mixed) & No \\
\midrule
\textbf{IMDD-1M (Ours)} & 2025 & \textbf{1.24M} & \textbf{63 (Diverse)} & \textbf{Yes (Image-Text Pairs)} \\
\bottomrule
\end{tabular}}
\vspace{-0.1in}
\end{table}

\section{Large-Scale Industrial Defect Dataset}
\label{sec:dataset}
\subsection{Data Collection}
\label{subsec:collection}
\hspace{1em}We consolidate large-scale data from public benchmarks, web mining, and industrial partnerships. Public datasets including DAGM~\cite{DAGM}, MVTec AD~\cite{MVTecAD}, KolektorSDD~\cite{KolektorSDD}, VisA~\cite{VisA}, and BTAD~\cite{BTAD} establish baseline coverage for canonical defect categories. We perform extensive web mining across GitHub, RoboFlow, PaddlePaddle, and Tianchi using multilingual queries in English, Chinese, and Japanese to capture diverse manufacturing products and defect terminologies.

We collaborate with industrial partners in petrochemical, metal processing, and powder metallurgy sectors to acquire authentic production-line imagery. Enterprise samples encompass polymer containers, chemical pipelines, castings, forgings, and sintered components exhibiting defects including corrosion, delamination, voids, inclusions, and surface pitting. All industrial data are anonymized. We implement stratified sampling ensuring balanced representation across product types and defect severities. The 18-month collection applies systematic quality control rejecting low-resolution or ambiguous samples.

\subsection{Annotation Framework}
\label{subsec:annotation}
\hspace{1em}Each image pairs with textual descriptions capturing product identity, defect morphology, and manufacturing causes. All annotations are performed by domain experts ensuring accurate capture of subtle defect characteristics and specialized terminology.
Figure \ref{fig:over} illustrates representative samples from the IMDD-1M dataset, showcasing the diversity of image-text pairs across industrial domains.

Industrial terminology is standardized across languages with a controlled vocabulary of over 500 specialized terms. Each description follows a structured template including product category, material composition, defect type, spatial location, and root causes. Annotations incorporate morphological descriptors including orientation (e.g., "vertical crack"), scale (e.g., "microscopic void"), localization (e.g., "upper-left corner"), and pattern (e.g., "radial distribution"). These attributes enable vision-language tasks such as root cause analysis and process optimization. This expert-driven approach offers greater precision in multimodal alignment compared to crowdsourcing.

\begin{figure}[t]
    \centering
    \includegraphics[width=1\columnwidth]{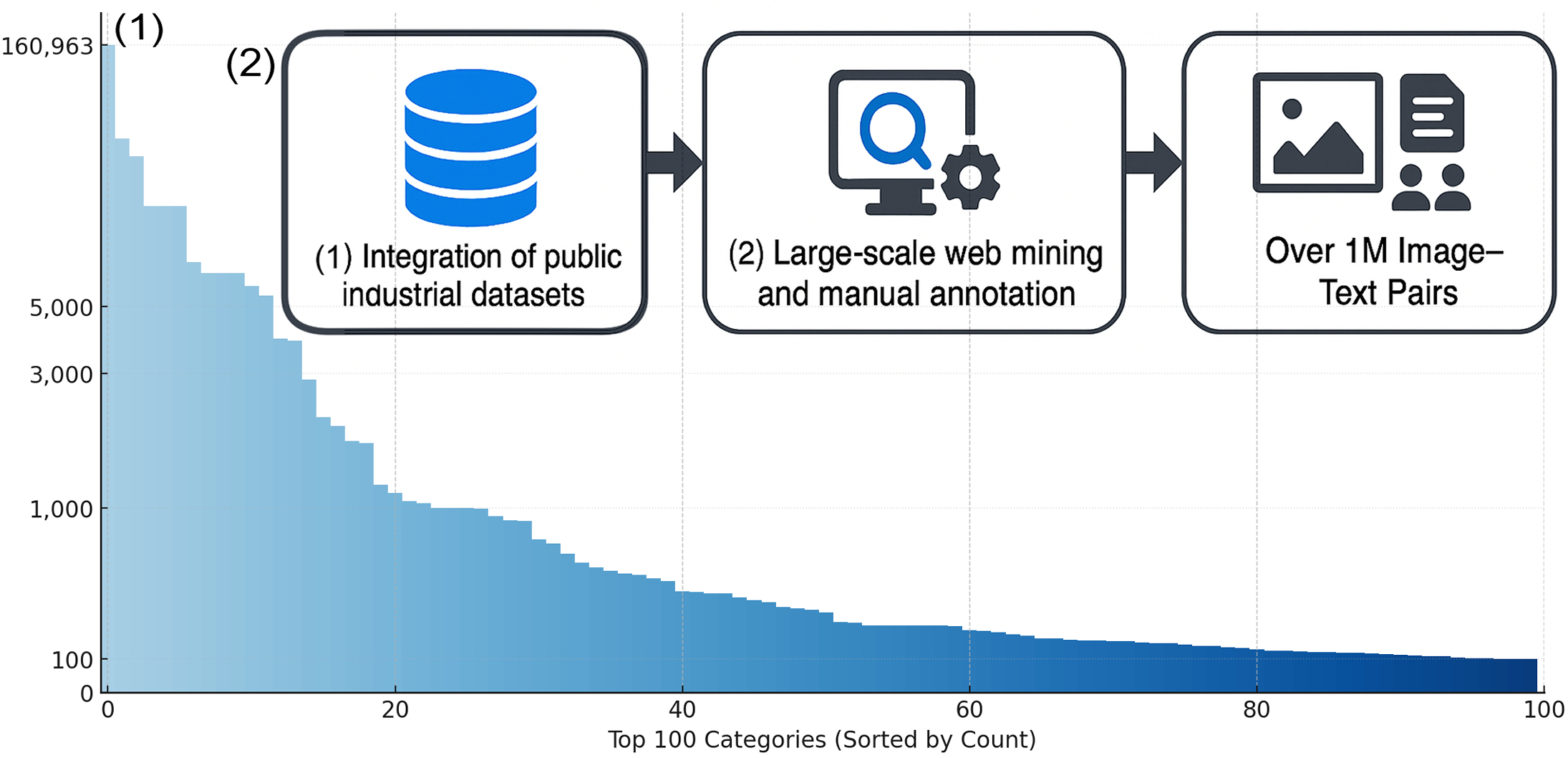}
    \vspace{-0.7em}
    \caption{
Dataset analysis. (1) Sample distribution among the top 100 defect categories (log-scaled). (2) Three-step workflow for dataset construction.
    }
    \label{fig:three}
\end{figure}

\begin{figure}[t]
    \centering
    \includegraphics[width=1\columnwidth]{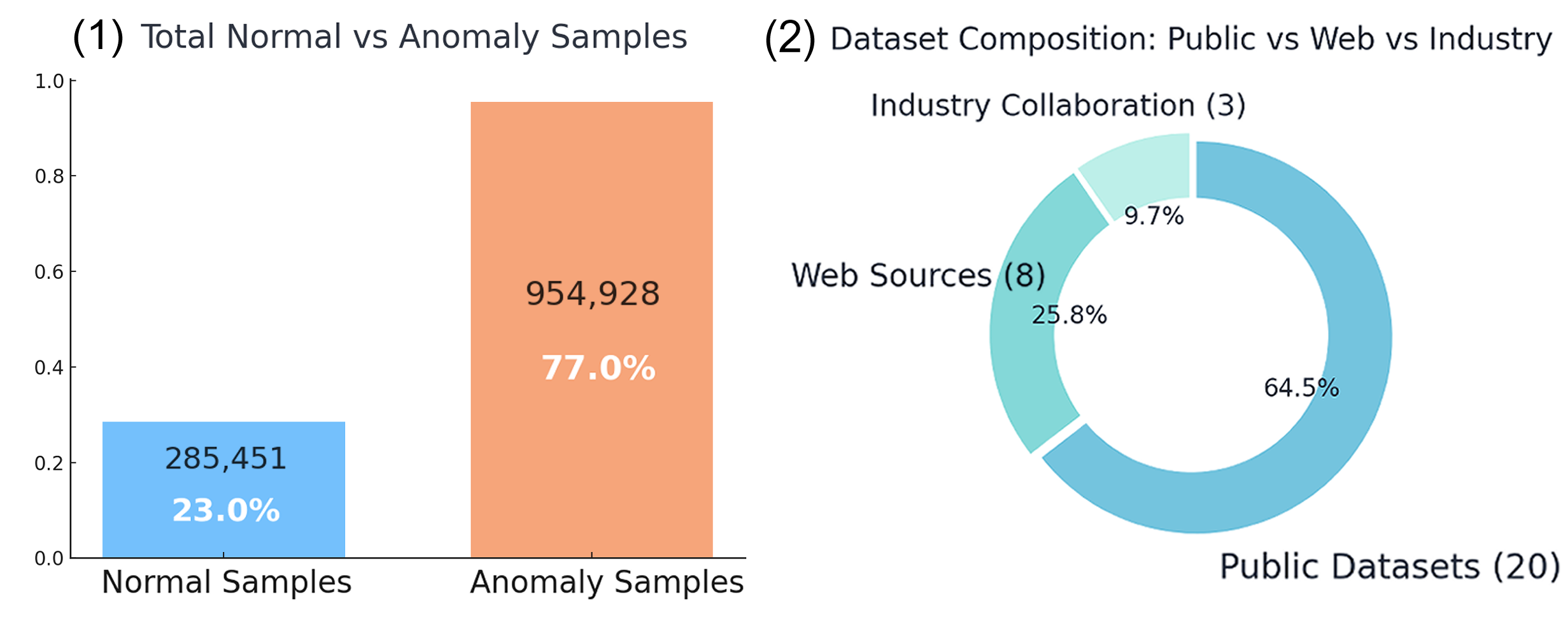}
    \vspace{-0.7em}
    \caption{
Dataset composition. (1) Distribution of normal versus anomaly samples. (2) Pie chart showing dataset composition across domains.
    }
    \label{fig:figg}
\end{figure}

\begin{table}[t]
\centering
\small
\caption{
Summary of Industrial Defect Datasets. 
Note: Enterprise collaboration data from petrochemical, metal processing, and powder metallurgy sectors are not included due to confidentiality.
}
\label{tab:dataset_summary}

\begin{adjustbox}{width=\columnwidth, center}
\begin{tabular}{lrrrrr}
\toprule
\textbf{Dataset} & \textbf{Cls.} & \textbf{Def. types} & \textbf{Normal} & \textbf{Anomaly} & \textbf{All} \\
\midrule
BTAD & 3 & 3 & 2250 & 290 & 2540 \\
SSGD & 1 & 7 & 0 & 2504 & 2504 \\
MVTec AD & 15 & 90 & 4096 & 1258 & 5354 \\
MVTec AD2 & 8 & 32 & 4705 & 3299 & 8004 \\
VisA & 12 & 137 & 9621 & 1200 & 10821 \\
Magnetic Tile & 1 & 6 & 952 & 392 & 1344 \\
NEU-DET & 1 & 7 & 0 & 1816 & 1816 \\
TIAN CHI Aluminum & 1 & 22 & 0 & 3311 & 3311 \\
TIAN CHI Fabric & 1 & 34 & 0 & 4762 & 4762 \\
TIAN CHI Bai Jiu & 2 & 14 & 1145 & 2225 & 3370 \\
SDI \& SPDI & 1 & 2 & 0 & 35 & 35 \\
Steel Pipe & 1 & 7 & 0 & 6966 & 6966 \\
WM-811K & 1 & 9 & 0 & 811457 & 811457 \\
ICCAD & 1 & 2 & 160963 & 4053 & 165016 \\
Tungsten Inert Gas & 1 & 6 & 30080 & 14978 & 45058 \\
Semiconductor & 1 & 2 & 27420 & 6696 & 34116 \\
Solar Cell & 2 & 8 & 1545 & 455 & 2000 \\
VOC2007 & 1 & 4 & 0 & 805 & 805 \\
Water Cooled & 1 & 1 & 0 & 321 & 321 \\
Aircraft Skin & 1 & 5 & 0 & 4281 & 4281 \\
OLED & 1 & 6 & 674 & 1854 & 2528 \\
Solar Cell Crack & 1 & 2 & 0 & 364 & 364 \\
Gear & 1 & 7 & 0 & 1719 & 1719 \\
PCB & 1 & 2 & 0 & 2100 & 2100 \\
Concrete & 1 & 2 & 20000 & 11688 & 31688 \\
Decks & 1 & 2 & 20000 & 10000 & 30000 \\
Wall & 1 & 2 & 2000 & 56099 & 58099 \\
\midrule
\textbf{Sum} & \textbf{63} & \textbf{421} & \textbf{285451} & \textbf{954928} & \textbf{1240379} \\
\bottomrule
\end{tabular}
\end{adjustbox}
\end{table}

\subsection{Dataset Statistics}
\label{subsec:statistics}
\begin{figure}[t]
    \centering
    \includegraphics[width=1\columnwidth]{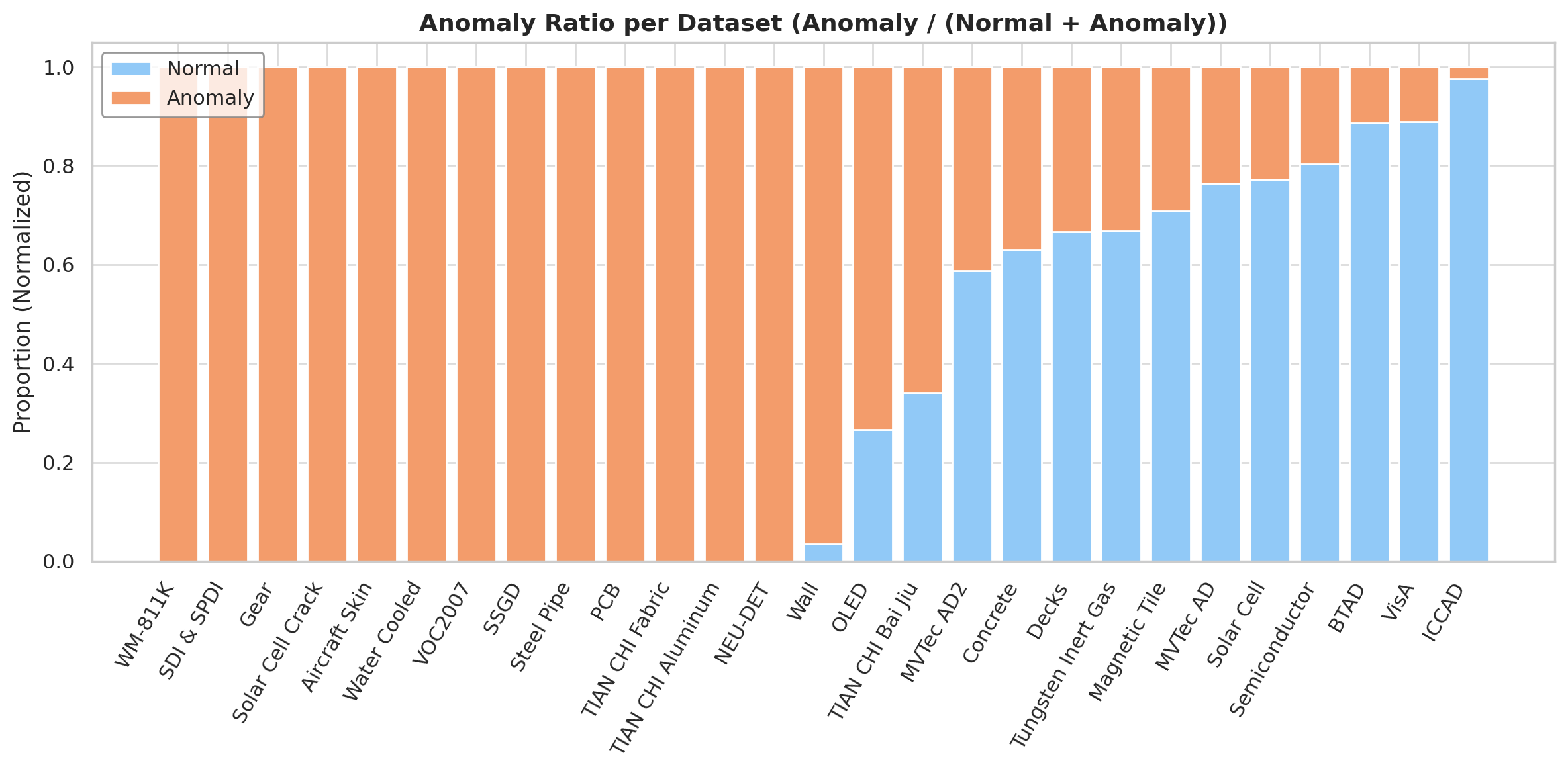}
    \vspace{-0.7em}
    \caption{
    Anomaly ratio distribution across datasets.
    Each bar represents the normalized proportion of anomaly and normal samples within a specific dataset, illustrating data imbalance and diversity across industrial domains.
    }
    \label{fig:figgg}
\end{figure}

\begin{figure}[t]
    \centering
    \includegraphics[width=1\columnwidth]{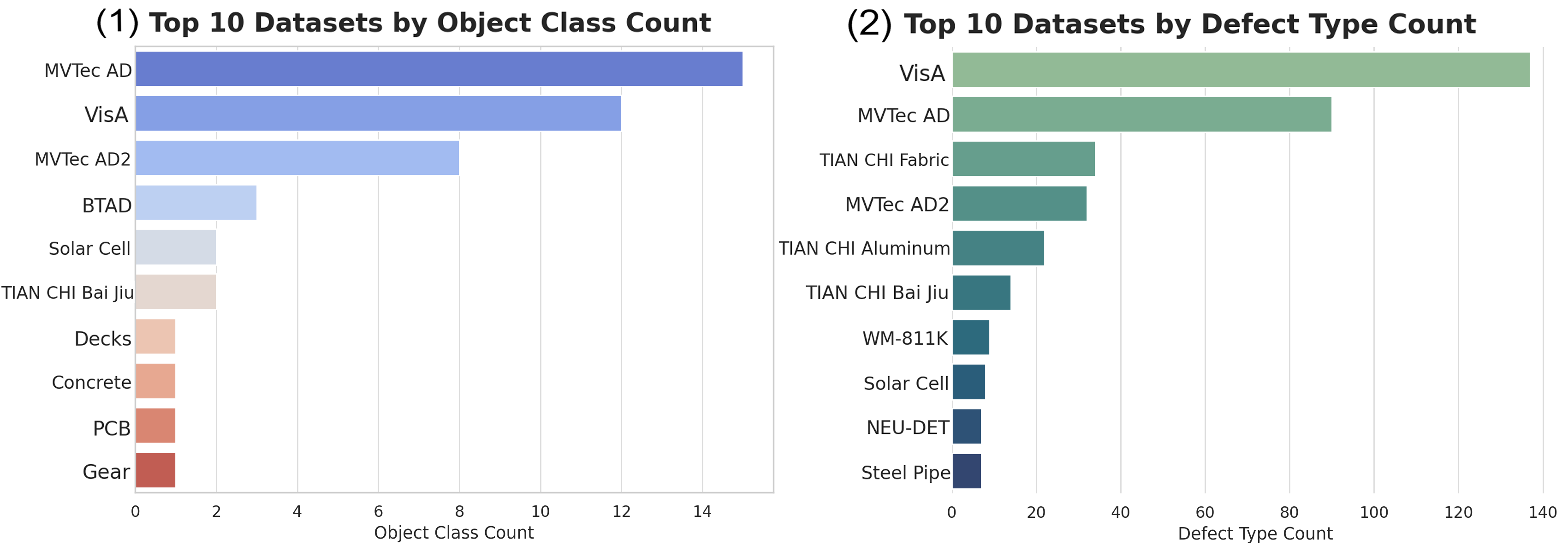}
    \vspace{-0.7em}
    \caption{
    Top 10 datasets ranked by (1) object class count and (2) defect type count.  
    MVTec AD and VisA stand out for their broad coverage and diversity across industrial components and surface conditions.
    }
    \label{fig:figggjj}
\end{figure}

\hspace{1em}The final dataset contains over 1.24 million high-resolution image-text pairs spanning 63 industrial product categories and 421 defect types. All images are standardized at 512×512 pixels, ensuring consistent input dimensions. Text descriptions average 42 words, providing rich semantic context without excessive verbosity.
Table~\ref{tab:dataset_summary} provides a comprehensive summary of incorporated datasets. The corpus comprises 285{,}451 normal and 954{,}928 anomaly samples, integrating diverse public benchmarks alongside large-scale industrial data. Figure~\ref{fig:three} illustrates proportional composition across product domains and defect categories, while Figure~\ref{fig:figg} presents the balance between normal and anomaly samples. Figure~\ref{fig:figgg} analyzes the anomaly ratio within each dataset, revealing substantial imbalance patterns across industrial domains. Figure~\ref{fig:figggjj} highlights the top 10 datasets ranked by object class and defect type count, showing datasets such as MVTec AD~\cite{MVTecAD} and VisA~\cite{VisA} offer the broadest coverage of categories and defect variations.
Industrial data samples introduce authentic variability, such as uneven illumination, complex backgrounds, and subtle defect manifestations, forming a foundational resource for industrial defect detection and VLMs development.

\begin{figure*}[t]
    \centering
    \includegraphics[width=\textwidth]{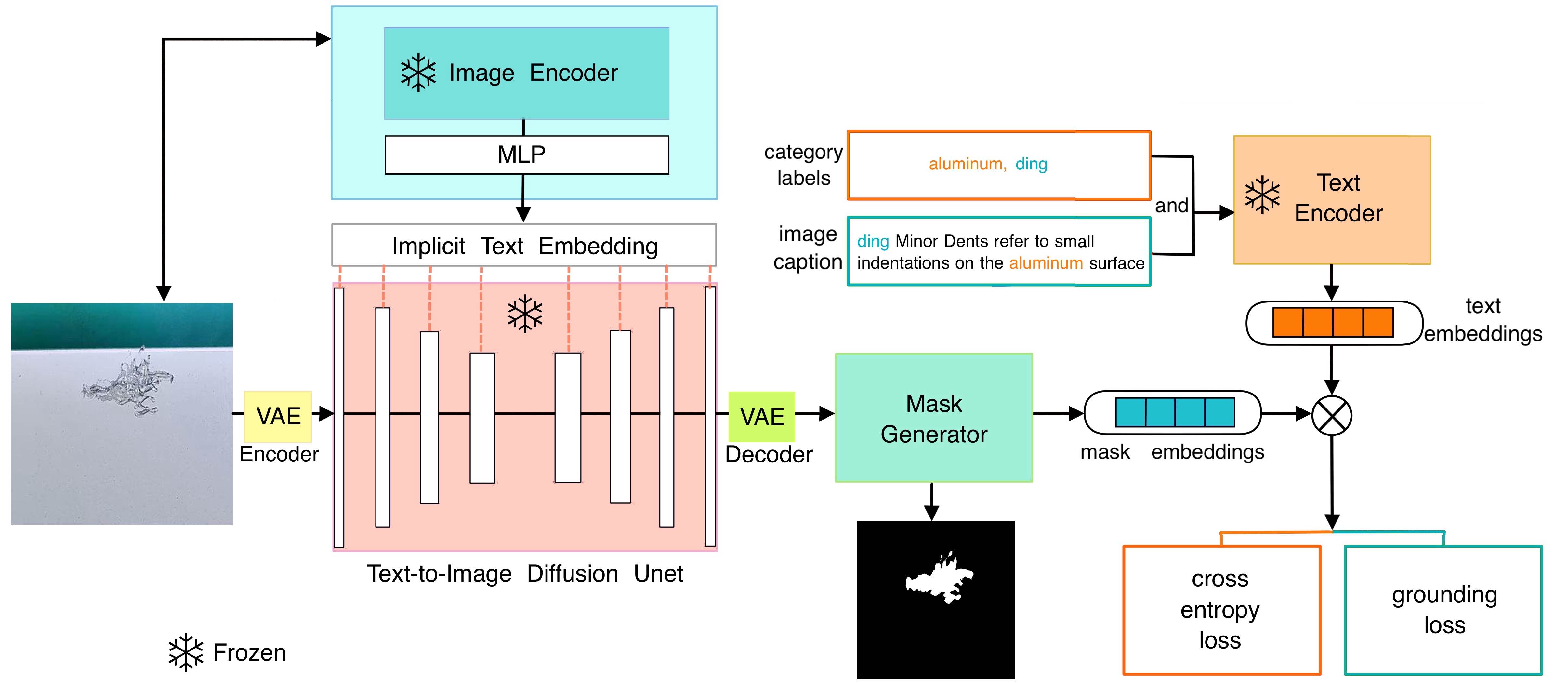}
    \vspace{-1em}
    \caption{Overview of our method. An implicit captioner encodes the defect image into a text embedding, which, together with the image, is fed into a frozen diffusion U-Net to extract multi-scale features. A VAE decoder reconstructs features, while a mask generator predicts binary masks and embeddings. Classification is performed via dot products between mask and text embeddings (orange/green boxes) under cross-entropy and grounding supervision.}
    \label{fig:overv}
\end{figure*}

\section{Method}
\label{sec:method}

\hspace{1em}We train a text-conditioned diffusion model from scratch on IMDD-1M, then transfer learned features to downstream tasks. The framework comprises an industrial diffusion U-shaped convolutional network (U-Net, 860M parameters), an implicit captioner (0.3M parameters), and a Mask2Former~\cite{mask2} generator (45M parameters).

\subsection{Problem Formulation}
\label{subsec:formulation}

\hspace{1em}Given image $\mathbf{I} \in \mathbb{R}^{H\times W\times 3}$ and optional text $\mathbf{t}$, we predict mask $\mathbf{M} \in \{0,1\}^{H\times W}$ with semantic label. We train on base categories $\mathcal{C}_{\text{train}}$ and test on disjoint $\mathcal{C}_{\text{test}}$ with only category names provided.

\subsection{Architecture}
\label{subsec:architecture}

\hspace{1em}We train a text-conditioned diffusion U-Net from scratch on IMDD-1M, where an implicit captioner generates pseudo text embeddings from visual features when captions are unavailable. The mask generator processes diffusion features to predict masks and embeddings for open-vocabulary classification.Figure \ref{fig:overv} illustrates the overall architecture of our proposed framework.

\subsection{Industrial Diffusion Model}
\label{subsec:diffusion_pretraining}

\hspace{1em}We adopt Stable Diffusion v1.5 U-Net with random initialization. The encoder has four blocks with channels 320, 640, 1280, 1280 at strides 1, 2, 4, 8. The decoder mirrors this with skip connections. Cross-attention layers inject text conditioning after every ResNet block. Given image $\mathbf{I}$, we encode via frozen Variational Autoencoder (VAE):
\begin{equation}
\mathbf{z}_0 = \mathcal{E}_{\text{VAE}}(\mathbf{I}) \in \mathbb{R}^{4 \times h \times w}, \quad h=H/8.
\end{equation}
\hspace{1em}We use frozen Stable Diffusion VAE for 8-fold compression. Noise follows Denoising Diffusion Probabilistic Models (DDPM):
\begin{equation}
\mathbf{z}_t = \sqrt{\bar{\alpha}_t}\mathbf{z}_0 + \sqrt{1-\bar{\alpha}_t}\boldsymbol{\epsilon}, \quad \boldsymbol{\epsilon} \sim \mathcal{N}(0, \mathbf{I}),
\end{equation}
with $\bar{\alpha}_t = \prod_{s=1}^t (1-\beta_s)$ and linear schedule $\beta_1=10^{-4}, \beta_T=0.02$ over $T=1000$ steps.

IMDD-1M images pair with captions like ``metal plate with scratches''. We encode via frozen CLIP:
\begin{equation}
\mathbf{e}_T = \text{CLIP}_{\text{text}}(\mathbf{t}) \in \mathbb{R}^{768}.
\end{equation}
\hspace{1em}The U-Net predicts noise conditioned on text via cross-attention:
\begin{equation}
\boldsymbol{\epsilon}_\theta(\mathbf{z}_t, t, \mathbf{e}_T) = \text{U-Net}_\theta(\mathbf{z}_t, t, \mathbf{e}_T).
\end{equation}
\hspace{1em}We minimize diffusion loss:
\begin{equation}
\mathcal{L}_{\text{diff}} = \mathbb{E}_{\mathbf{z}_0, \boldsymbol{\epsilon}, t} \left[ \|\boldsymbol{\epsilon} - \boldsymbol{\epsilon}_\theta(\mathbf{z}_t, t, \mathbf{e}_T)\|_2^2 \right],
\end{equation}
where $t \sim \text{Uniform}(1, T)$. Training runs 100 epochs on 1,240,379 images with batch size 256 across 8 H100 GPUs, requiring 72 hours. All 860M parameters train from random initialization.

\subsection{Implicit Captioner}
\label{subsec:implicit}

\hspace{1em}Downstream datasets typically lack captions, providing only categorical labels or binary normal-versus-defective annotations. This poses a challenge for extracting diffusion features which require text conditioning. We address this by introducing an implicit captioner that synthesizes pseudo text embeddings directly from images, eliminating the need for explicit captions during both training and inference. The module consists of a frozen CLIP image encoder followed by a trainable two-layer MLP projecting 512-dimensional CLIP embeddings into the 768-dimensional text embedding space:
\begin{equation}
\resizebox{!}{6.5pt}{$
\mathbf{t}_{\text{imp}} = \text{MLP}_\phi(\mathcal{V}(\mathbf{I})) = \mathbf{W}_2 \cdot \text{GELU}(\mathbf{W}_1 \cdot \mathcal{V}(\mathbf{I}) + \mathbf{b}_1) + \mathbf{b}_2
$}
\end{equation}

During Stage 1 pretraining, we train the implicit captioner jointly with the diffusion U-Net via stochastic conditioning. For each training sample, we randomly select whether to condition on ground-truth captions or implicit embeddings with equal probability:
\begin{equation}
\mathbf{c} \sim \begin{cases}
\mathbf{e}_T & \text{prob. } 0.5 \\
\mathbf{t}_{\text{imp}} & \text{prob. } 0.5
\end{cases}
\end{equation}
\hspace{1em}This training strategy encourages the implicit embeddings to serve as effective substitutes for real text embeddings. We further enforce alignment through an auxiliary cosine similarity loss:
\begin{equation}
\mathcal{L}_{\text{imp}} = 1 - \frac{\mathbf{t}_{\text{imp}}^T \mathbf{e}_T}{\|\mathbf{t}_{\text{imp}}\| \|\mathbf{e}_T\|}.
\end{equation}

\subsection{Feature Extraction}
\label{subsec:feature_extraction}

\hspace{1em}After pretraining, we freeze the diffusion model and extract features via single forward pass. Given image $\mathbf{I}$, we add noise at $t=50$ providing optimal semantic-spatial balance:
\begin{equation}
\mathbf{I}_t = \sqrt{\bar{\alpha}_{50}}\mathbf{I} + \sqrt{1-\bar{\alpha}_{50}}\boldsymbol{\epsilon}.
\end{equation}
\hspace{1em}We encode to latent $\mathbf{z}_t = \mathcal{E}_{\text{VAE}}(\mathbf{I}_t)$, generate implicit caption $\mathbf{t}_{\text{imp}} = \text{MLP}_\phi(\mathcal{V}(\mathbf{I}))$, and extract features:
\begin{equation}
\{\mathbf{f}_\ell\}_{\ell=1}^4 = \text{U-Net}_\theta(\mathbf{z}_t, 50, \mathbf{t}_{\text{imp}}).
\end{equation}
\hspace{1em}Features have resolutions $\{h, h/2, h/4, h/8\}$ with channels $\{320, 640, 1280, 1280\}$.

\subsection{Mask Generation and Classification}
\label{subsec:downstream}

\hspace{1em}The mask generator adopts Mask2Former with pixel decoder and transformer decoder. The pixel decoder implements Feature Pyramid Network (FPN) producing $\mathbf{F} \in \mathbb{R}^{256 \times h \times w}$. The transformer decoder uses 100 learnable queries attending to pixel features, producing masks $\{\mathbf{m}_i\}_{i=1}^{100}$ and embeddings $\{\mathbf{z}_i\}_{i=1}^{100}$. We supervise with binary cross-entropy:
\begin{equation}
\resizebox{!}{6.5pt}{$
\mathcal{L}_{\text{mask}} = -\sum_{i,j} \left[\mathbf{M}_{ij}\log \mathbf{m}_{ij} + (1 - \mathbf{M}_{ij})\log(1 - \mathbf{m}_{ij})\right].
$}
\end{equation}

For classification, with category labels we encode training categories via CLIP forming $\mathbf{T} = [\text{CLIP}_{\text{text}}(c_1), \ldots, \text{CLIP}_{\text{text}}(c_K)]$. For mask embedding $\mathbf{z}_i$ with label $y_i$:
\begin{equation}
\mathcal{L}_{\text{cls}} = \frac{1}{N}\sum_{i=1}^N \text{CE}\left(\text{Softmax}(\mathbf{z}_i \cdot \mathbf{T}^T / \tau), y_i\right).
\end{equation}

With captions only, we extract nouns as pseudo-labels. Given batch $\{(\mathbf{I}^{(m)}, s^{(m)})\}_{m=1}^B$ with nouns $\mathcal{C}_{\text{word}}^{(m)}$, we compute grounding similarity:
\begin{equation}
\resizebox{!}{8.4pt}{$
g(\mathbf{I}^{(m)}, s^{(m)}) =
\frac{1}{K_w}\sum_{k=1}^{K_w} \sum_{i=1}^N
p(\mathbf{z}_i)_k \cdot
\langle \mathbf{z}_i, \text{CLIP}_{\text{text}}(w_k) \rangle
$}
\end{equation}
where $p(\mathbf{z}_i)_k$ denotes the $k$-th element of the softmax-normalized similarity. We apply bidirectional contrastive loss:
\begin{equation}
\resizebox{!}{31pt}{$
\begin{split}
\mathcal{L}_{\text{ground}} = -\frac{1}{B}\sum_{m=1}^B \Bigg[
& \log \frac{\exp(g(\mathbf{I}^{(m)}, s^{(m)})/\tau)}{\sum_{n=1}^B \exp(g(\mathbf{I}^{(m)}, s^{(n)})/\tau)} \\
& + \log \frac{\exp(g(\mathbf{I}^{(m)}, s^{(m)})/\tau)}{\sum_{n=1}^B \exp(g(\mathbf{I}^{(n)}, s^{(m)})/\tau)}
\Bigg].
\end{split}
$}
\end{equation}

\subsection{Training Protocol}
\label{subsec:training}

\hspace{1em}Stage 1 trains entire diffusion model from scratch on IMDD-1M for 100 epochs:
\begin{equation}
\mathcal{L}_{\text{Stage1}} = \mathcal{L}_{\text{diff}} + 0.3 \mathcal{L}_{\text{imp}}.
\end{equation}
\hspace{1em}All parameters train: U-Net (860M) and implicit captioner (0.3M). AdamW optimizer, learning rate $1 \times 10^{-4}$, batch size 256, 72 hours on 8 H100 GPUs.

Stage 2 freezes diffusion model, trains mask generator on downstream datasets for 50 epochs:
\begin{equation}
\mathcal{L}_{\text{Stage2}} = \mathcal{L}_{\text{mask}} + 0.5 \mathcal{L}_{\text{cls/ground}}.
\end{equation}
\hspace{1em}Mask generator (45M) trains with AdamW, learning rate $5 \times 10^{-5}$, batch size 16, 4 hours on 8 H100 GPUs.

At test time with novel categories $\mathcal{C}_{\text{test}}$, we extract features, generate masks and embeddings, and classify via $\hat{y}_i = \arg\max_c p(\mathbf{z}_i, \mathcal{C}_{\text{test}})_c$. Inference requires 0.35s per image on A100 GPU.

\section{Experiments}
\label{sec:experiments}
\subsection{Implementation Details}
\label{subsec:implementation}
\hspace{1em}We provide complete architecture, training, and evaluation details in the supplementary material. Our model comprises 890M parameters with 0.35s inference time per image on an A100 GPU. 
We evaluate generative quality using the Fr\'echet Inception Distance (FID) and Inception Score (IS), computed between generated and real defect samples.

\subsection{Text-Guided Defect Generation}
\label{subsec:generation}

\hspace{1em}After training on IMDD-1M, our model generates realistic defect patterns from textual descriptions such as ``bottle with contamination'' or ``metal surface with oxidation.'' This demonstrates meaningful multimodal representations learned through large-scale training.

\begin{figure}[t]
\centering
\setlength{\abovecaptionskip}{4pt}
\setlength{\belowcaptionskip}{-4pt}

\includegraphics[width=1\linewidth]{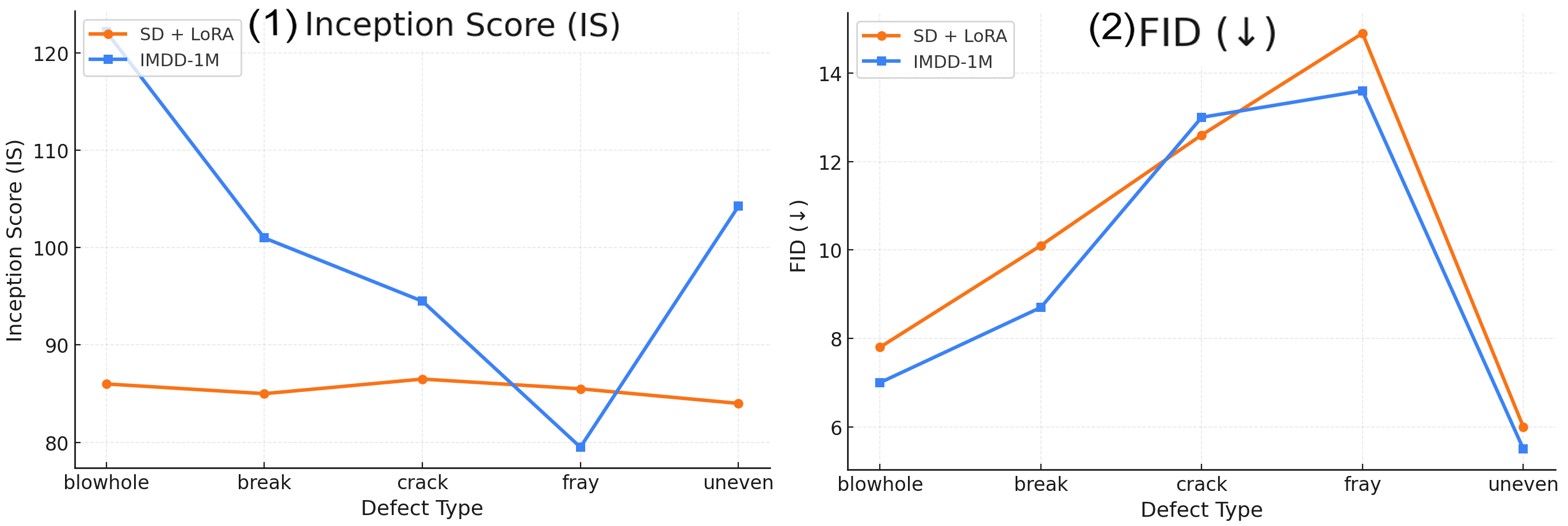}

\caption{
Comparison of generative quality between our IMDD-1M–trained model and Stable Diffusion XL (SDXL) on the Magnetic Tile dataset.  
(1) IS: class consistency and diversity.   
(2) FID: realism gap to real images.   
Our model attains higher IS and lower FID.
}
\label{fig:is_fid}
\end{figure}

\begin{figure}[t]
\centering
\includegraphics[width=\linewidth]{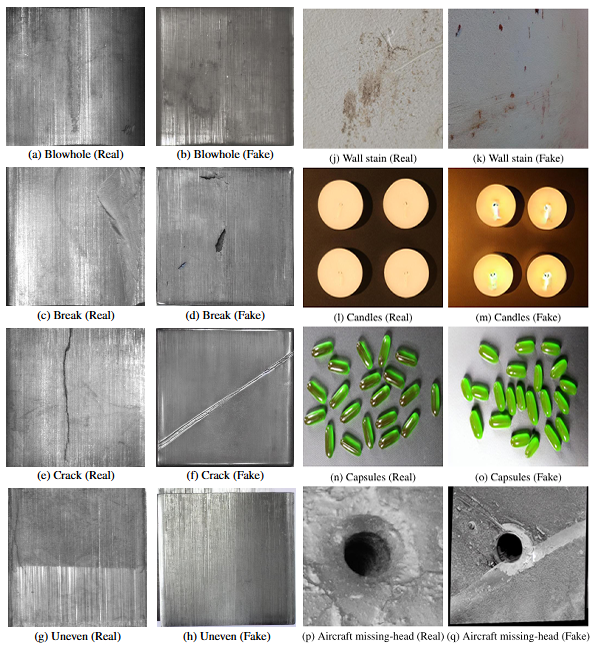}
\caption{Qualitative comparison of real (left) vs. generated (right) defect samples across multiple industrial datasets including Magnetic Tile, VisA, wall stain, and aircraft surface panel. Generated images exhibit high fidelity in texture reproduction.}
\label{fig:generation}
\end{figure}

Figure~\ref{fig:is_fid} shows our model achieves 100.29 IS and 5.5-13.6 FID using IMDD-1M-trained features, confirming realistic and diverse generation. Figure~\ref{fig:generation} demonstrates synthesized images preserve material-specific visual characteristics where metallic surfaces show appropriate reflectance while textile defects maintain fiber structure. These generated samples provide controllable synthetic augmentation for downstream tasks, expanding training distributions to improve robustness for rare defect types. Captioning evaluation is deferred to the future work.

\begin{figure*}[t]
    \centering
    \includegraphics[width=\textwidth]{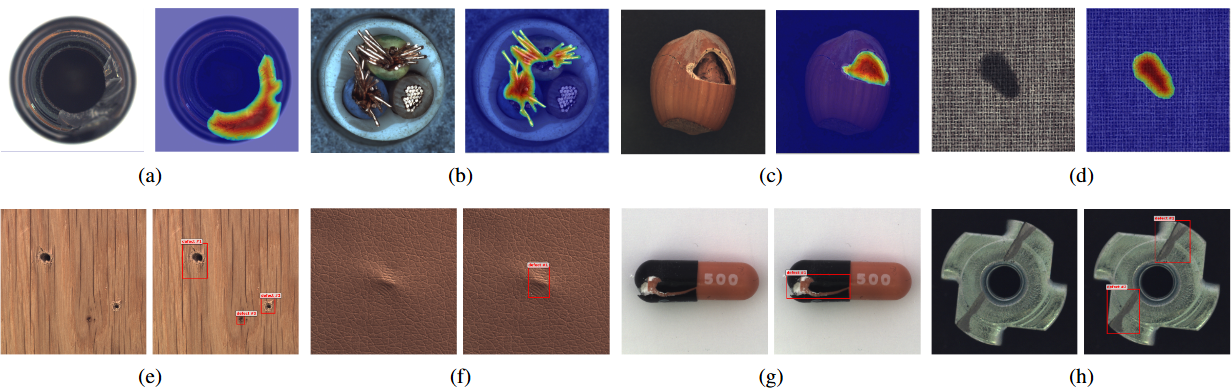}
    \vspace{-1em}
    \caption{
    Qualitative visualization of multimodal results on various MVTec AD samples. 
    (a)–(d) show segmentation outputs with text-conditioned masks highlighting localized defects, 
    while (e)–(h) illustrate object detection results with bounding boxes accurately identifying defect regions across different material domains.
    }
    \label{fig:overview}
\end{figure*}
\subsection{Unified Framework for Downstream Tasks}
\label{subsec:downstream}

\hspace{1em}We evaluate on classification, object detection, and segmentation tasks to demonstrate effective transfer of learned representations.

\subsubsection{Defect Classification}

\begin{table}[t]
\centering 
\caption{Classification accuracy across multiple industrial datasets.}
\label{tab:classification}
\begin{tabular*}{\columnwidth}{@{\extracolsep{\fill}} lcc @{}}
\toprule
\textbf{Dataset} & \textbf{\# Defect Types} & \textbf{Accuracy (\%)} \\
\midrule
MVTec AD & 90 & 98.3 \\
VisA & 137 & 97.7 \\
Magnetic Tile & 6 & 96.2 \\
Steel Surface & 7 & 94.5 \\
\midrule
\textbf{Average} & -- & \textbf{96.7} \\
\bottomrule
\end{tabular*}
\vspace{-0.1in}
\end{table}

\hspace{1em}We show classification results in Table~\ref{tab:classification}. Our model achieves 96.7 \% average accuracy across four datasets without task-specific modifications.

\subsubsection{Object-Level Defect Detection}

\begin{table}[t]
\centering
\caption{
Object detection comparison with YOLOv8 on MVTec AD. 
We report the mean Average Precision (mAP) at Intersection over Union (IoU) thresholds of 0.5 and 0.75, and the average IoU.
}
\label{tab:detection}
\begin{adjustbox}{width=\columnwidth, center}
\begin{tabular}{lccc}
\toprule
\textbf{Method} & \textbf{mAP@0.5 (\%)} & \textbf{mAP@0.75 (\%)} & \textbf{Avg IoU (\%)} \\
\midrule
\textbf{YOLOv8-m}~\cite{yolov8} & 78.3 & 62.1 & 68.4 \\
Ours (Mask-based) & 74.6 & 58.9 & 65.2 \\
\bottomrule
\end{tabular}
\end{adjustbox}
\vspace{-0.05in}
\end{table}

\hspace{1em}We derive bounding boxes from segmentation masks. Table~\ref{tab:detection} shows that our mask-based defect foundation model, fine-tuned with only 200 samples per class (less than 5\% of data required by supervised methods), achieves 74.6\% mAP@0.5 and 58.9\% mAP@0.75, approaching the performance of the dedicated object detection model YOLOv8 (78.3\% and 62.1\% respectively). This demonstrates exceptional data efficiency and strong generalization without requiring explicit box annotations, as our unified framework rivals task-specific detectors while using significantly fewer labeled samples.

\subsubsection{Pixel-Level Defect Segmentation}

\begin{table}[t]
\centering
\caption{
Pixel-level segmentation results using standard metrics: Accuracy, F1-score (F1), IoU, and Dice coefficient (Dice), evaluated on ground-truth masks from MVTec AD and VisA.
}
\label{tab:segmentation}

\begin{adjustbox}{width=\columnwidth, center}
\begin{tabular}{lcccc}
\toprule
\textbf{Dataset} & \textbf{Accuracy (\%)} & \textbf{F1 (\%)} & \textbf{IoU (\%)} & \textbf{Dice (\%)} \\
\midrule
MVTec AD (bottle) & 92.25 & 58.3 & 52.1 & 58.3 \\
MVTec AD (cable) & 89.7 & 56.8 & 51.4 & 56.8 \\
VisA (candle) & 90.3 & 60.2 & 54.7 & 60.2 \\
VisA (capsule) & 91.8 & 59.1 & 53.3 & 59.1 \\
\midrule
\textbf{Average} & \textbf{91.0} & \textbf{58.6} & \textbf{52.9} & \textbf{58.6} \\
\bottomrule
\end{tabular}
\end{adjustbox}

\vspace{-0.1in}
\end{table}

\begin{table*}[t]
\centering
\caption{Comparison with anomaly detection methods on MVTec AD dataset. We report P-AUC-ROC (\%): area under the receiver operating characteristic curve at pixel level, and AUC-PRO (\%): area under the per-region overlap curve. Our method achieves competitive performance with significantly reduced supervision (200 samples/class vs. full (4000 after data augmentation) training sets).}
\label{tab:mvtec_comparison}
\resizebox{\textwidth}{!}{
\begin{tabular}{l|ccccccc}
\toprule
\textbf{Method} &
\textbf{MuSc~\cite{musc}} &
\textbf{PromptAD~\cite{promptad}} &
\textbf{DMAD~\cite{dmad}} &
\textbf{DDAD~\cite{ddad}} &
\textbf{SimpleNet~\cite{simplenet}} &
\textbf{FAIR~\cite{fair}} &
\textbf{Ours (50 samples/class)} \\
\midrule
P-AUC-ROC &
97.3 &
96.5 &
97.9 &
98.1 &
98.1 &
\textbf{98.2} &
96.1 \\
AUC-PRO &
93.8 &
90.5 &
93.3 &
92.3 &
90.5 &
\textbf{94.0} &
90.2 \\
\midrule
\# Training Samples & Full & Full & Full & Full & Full & Full & \textbf{200/class} \\
\bottomrule
\end{tabular}}
\vspace{-0.1in}
\end{table*}

\hspace{1em}We report segmentation results in Table~\ref{tab:segmentation} achieving 52.9 \% average IoU. Table~\ref{tab:mvtec_comparison} compares with existing methods on MVTec AD. Our approach with 200 samples per class achieves 96.1 \% P-AUC-ROC and 90.2 \% AUC-PRO, approximately 2 \% below methods using full training sets. This modest gap is favorable considering we use less than 5\% of the annotation requirements of supervised methods. As shown in Figure~\ref{fig:overview}, our multimodal model further provides visually interpretable segmentation and detection outputs, effectively localizing diverse defect patterns across different material domains.

\subsection{Ablation Study and Data Efficiency}
\label{subsec:ablation}

\begin{table}[t]
\centering
\caption{Architectural ablation study on VisA dataset. Removing the implicit text embedding, grounding loss, or diffusion conditioning leads to consistent degradation, confirming that each component contributes to overall accuracy and segmentation quality.}
\label{tab:ablation_arch}
\resizebox{\columnwidth}{!}{
\begin{tabular}{l|c|c|c}
\toprule
\textbf{Model Configuration} & \textbf{Acc (\%)} & \textbf{F1 (\%)} & \textbf{IoU (\%)} \\
\midrule
Full Model & \textbf{91.0} & \textbf{58.6} & \textbf{52.9} \\
w/o Implicit Text Embedding & 86.2 & 54.1 & 49.2 \\
w/o Grounding Loss $\mathcal{L}_{\text{grnd}}$ & 88.3 & 56.4 & 49.8 \\
w/o Diffusion Conditioning & 84.0 & 52.3 & 46.7 \\
\bottomrule
\end{tabular}}
\end{table}

\hspace{1em}Table~\ref{tab:ablation_arch} evaluates component contributions. Removing implicit text embedding reduces accuracy by 4.8 \%, eliminating grounding loss degrades IoU by 3.1 \%, and removing diffusion conditioning causes 7.0 \% accuracy drop.

\begin{figure}[t]
\centering
\includegraphics[width=\linewidth]{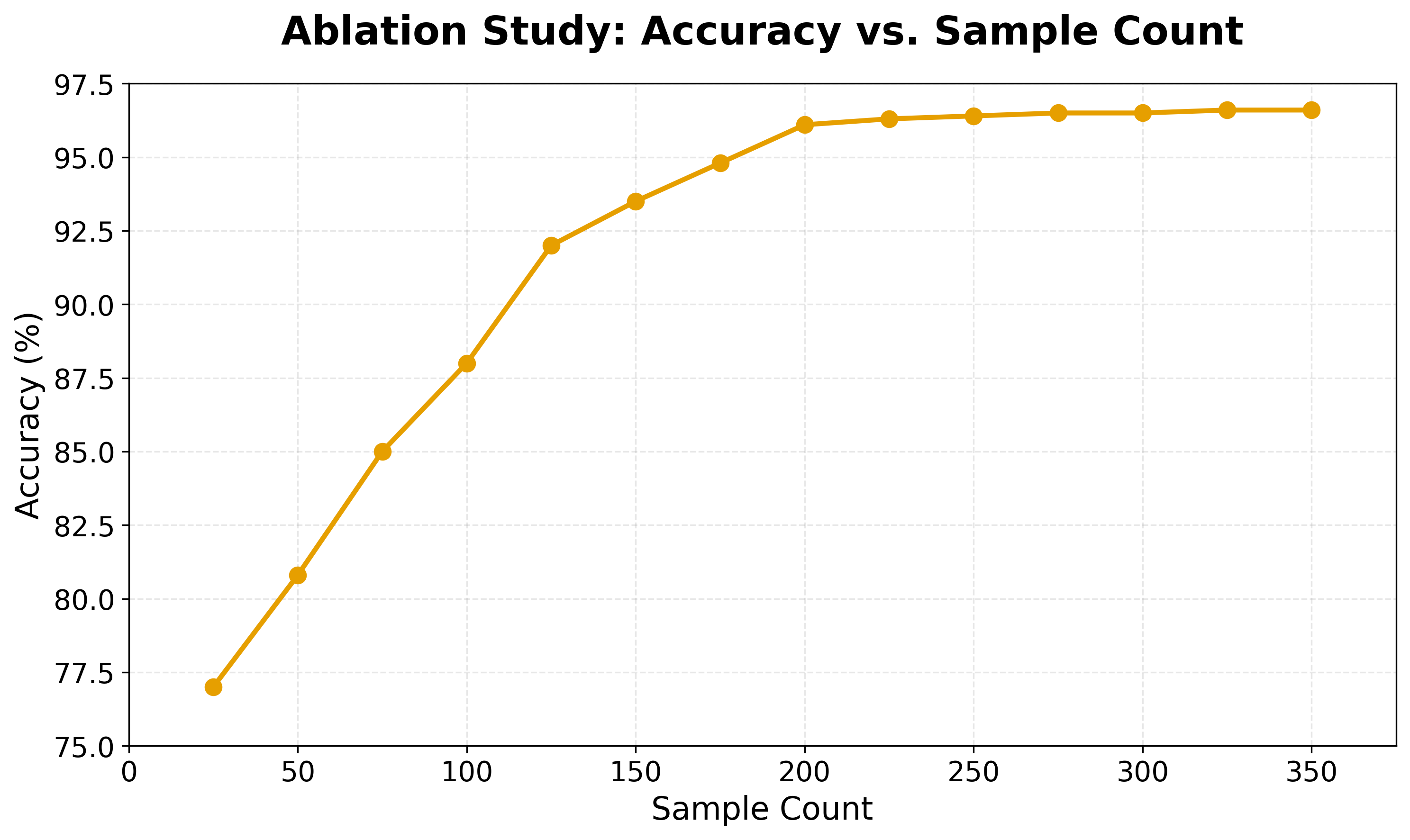}
\caption{
Our method achieves 96.1\% accuracy using only 200 samples per class, requiring less than 5\% of the training data compared to conventional approaches (approximately 4,000 samples per class for comparable performance). 
Performance rapidly improves up to 200 samples and then plateaus, demonstrating effective learning under limited supervision.
}

\label{fig:data_efficiency}
\end{figure}

Figure~\ref{fig:data_efficiency} demonstrates the data efficiency of our approach through an ablation study varying training samples from 25 to 350 per class.
Our model, pre-trained on the large-scale IMDD-1M dataset, achieves 96.1\% accuracy with only 200 samples per class during fine-tuning.

In contrast, conventional supervised methods typically require approximately 4,000 samples per class (including augmentation) to reach comparable performance, meaning our approach reduces annotation requirements to less than 5\% while maintaining competitive accuracy.
The performance curve shows rapid improvement in the low-data regime (25-200 samples) and saturates beyond 200 samples, indicating that our foundation model has learned generalizable defect representations.

\section{Conclusion}
\label{sec:conclusion}
We introduced IMDD-1M, a million-scale industrial defect dataset with aligned image-text annotations. 
Our unified diffusion framework achieves competitive zero-shot performance using less than 5\% of supervised training data while eliminating task-specific models. 
This work establishes a foundation for scalable, language-driven industrial inspection systems. 
In the future, we aim to extend the dataset with temporal and multi-view information to support video-based defect tracking and 3D reasoning. 
We also plan to explore cross-domain generalization between different manufacturing sectors, enabling robust adaptation to unseen industrial settings. 
Integrating multimodal reasoning with physical simulation will further bridge perception and generative modeling for real-world manufacturing intelligence.

{
\small 
\bibliographystyle{ieeenat_fullname} 
\bibliography{main}
}
\section*{Appendix}

\section{Reproducibility and Code Release}

To facilitate reproducibility and future research, we have released comprehensive code, lightweight pre-trained model snapshots, and detailed documentation through our GitHub repository:  
\url{https://github.com/NinaNeon/IMDD-1M-Towards-Open-Vocabulary-Industrial-Defect-}.  
This section provides an overview of the released materials and instructions for reproducing our experimental results.

\subsection{ Repository Structure}

Our codebase is organized into the following components:

\begin{itemize}
    \item \texttt{models/}: Core model implementations, including the Industrial Diffusion U-Net, Implicit Captioner, and Mask2Former generator.
    \item \texttt{third\_party/}: Integration of external third-party libraries used by our framework.
    \item \texttt{Object\_detection.py}: Object-level defect detection implementation.
    \item \texttt{classify.py}: Defect classification module.
    \item \texttt{integrate\_custom\_unet.py}: Utilities for integrating custom U-Net variants.
    \item \texttt{README.md}: Comprehensive documentation including installation, usage, examples, and training instructions.
    \item \texttt{requirements.txt}: Full dependency specification to reproduce all experiments.
    \item \texttt{LICENSE}: Open-source license governing code usage and redistribution.
\end{itemize}

\subsection{Model and Dataset Release}

Due to storage and hosting limits, we cannot release the full pre-trained models and datasets in the repository, but provide lightweight snapshots, configurations, and preparation scripts, with complete resources available to qualified researchers upon request under institutional data-sharing policies.

\section{Limitations}
\label{subsec:limitations}

\subsection{Training Limitations.}
While our IMDD-1M dataset comprises 1.24 million samples, training the diffusion U-Net from scratch requires substantial computational resources. The complete Stage 1 pre-training demands 72 hours on 8× NVIDIA H100 80GB GPUs (576 GPU-hours total), which may limit accessibility for researchers with constrained computational budgets. The peak memory consumption reaches 76GB per GPU at batch size 32 with mixed precision training, necessitating high-end hardware.

The two-stage training paradigm (diffusion pre-training followed by mask generator fine-tuning) introduces additional complexity compared to end-to-end approaches. Researchers must carefully manage frozen and trainable parameters across stages, and hyperparameter tuning requires iterating through both stages, multiplying computational costs.

\subsection{Inference Limitations.}
Our model requires 0.35 seconds per image on an A100 GPU, which may be slower than specialized detectors like YOLOv8 for real-time industrial inspection scenarios requiring 50-200 frames per second. The diffusion-based feature extraction at timestep $t=50$ adds computational overhead compared to standard feed-forward architectures. Memory consumption of 18.7GB during inference exceeds the capacity of edge devices commonly used in industrial settings.

\subsection{Application Limitations.}
Despite achieving competitive performance with less than 5\% of supervised training data (200 samples per class), our approach still requires this minimum amount for effective fine-tuning. For extremely rare defects occurring less than once per 10,000 products, collecting 200 samples may be impractical. The framework currently focuses on 2D analysis and does not incorporate temporal information for video-based inspection or 3D geometric reasoning for volumetric defects. IMDD-1M predominantly covers visible-light RGB imaging, while industrial settings often employ X-ray, infrared, ultrasonic, or hyperspectral imaging.

\section{Societal Impact}
\label{subsec:societal_impact}

\subsection{Positive Impacts.}
This work contributes to improved manufacturing quality control, potentially reducing defective products and enhancing consumer safety. By enabling data-efficient defect detection with reduced annotation requirements (less than 5\%), our approach democratizes access to advanced AI-powered inspection for small and medium-sized enterprises that lack extensive labeled datasets. The multimodal framework facilitates knowledge transfer across manufacturing domains, accelerating AOI adoption. Automated systems improve workplace safety by reducing human exposure to hazardous environments including high-temperature processes, toxic materials, and repetitive strain injuries.

\subsection{Potential Concerns.}
Deployment of automated defect detection systems may impact employment in traditional quality inspection roles, necessitating workforce retraining and transition support. There exist risks of automation bias where operators over-rely on AI predictions without verification. We emphasize human-in-the-loop workflows for safety-critical applications. The dataset contains proprietary manufacturing patterns; organizations should evaluate intellectual property concerns before releasing defect imagery. The computer vision techniques could be repurposed for surveillance or discriminatory practices. We advocate for responsible AI principles and regulatory frameworks preventing misuse.
\section{Preliminaries}

\subsection{Denoising Diffusion Probabilistic Models}

\subsubsection{Forward Diffusion Process}

Progressive noise addition over $T$ timesteps:
\begin{equation}
q(\mathbf{x}_t | \mathbf{x}_{t-1}) = \mathcal{N}(\mathbf{x}_t; \sqrt{1-\beta_t}\mathbf{x}_{t-1}, \beta_t\mathbf{I}),
\end{equation}
where $\{\beta_t\}_{t=1}^T$ controls noise injection rate.

\paragraph{Closed-Form Sampling.}
Define $\alpha_t = 1 - \beta_t$ and $\bar{\alpha}_t = \prod_{s=1}^t \alpha_s$. Through recursive substitution:
\begin{equation}
\mathbf{x}_t = \sqrt{\bar{\alpha}_t}\mathbf{x}_0 + \sqrt{1-\bar{\alpha}_t}\boldsymbol{\epsilon}, \quad \boldsymbol{\epsilon} \sim \mathcal{N}(\mathbf{0}, \mathbf{I}).
\end{equation}

This reparameterization enables efficient training without iterating through timesteps. As $t \to T$, we have $\mathbf{x}_T \approx \mathcal{N}(\mathbf{0}, \mathbf{I})$.

\paragraph{Variance Schedule.}
We use linear schedule: $\beta_t = \beta_1 + \frac{t-1}{T-1}(\beta_T - \beta_1)$ with $\beta_1 = 10^{-4}, \beta_T = 0.02$. Alternative cosine schedule: $\bar{\alpha}_t = \frac{f(t)}{f(0)}$ where $f(t) = \cos\left(\frac{t/T + s}{1+s} \cdot \frac{\pi}{2}\right)^2$ with $s=0.008$.

\subsubsection{Reverse Denoising Process}

Learn reverse process $p_\theta(\mathbf{x}_{0:T}) = p(\mathbf{x}_T) \prod_{t=1}^T p_\theta(\mathbf{x}_{t-1} | \mathbf{x}_t)$ where $p(\mathbf{x}_T) = \mathcal{N}(\mathbf{0}, \mathbf{I})$.

\paragraph{Posterior Distribution.}
The true reverse transition is:
\begin{align}
q(\mathbf{x}_{t-1} | \mathbf{x}_t, \mathbf{x}_0) &= \mathcal{N}(\mathbf{x}_{t-1}; \tilde{\boldsymbol{\mu}}_t, \tilde{\beta}_t\mathbf{I}), \\
\tilde{\boldsymbol{\mu}}_t &= \frac{\sqrt{\bar{\alpha}_{t-1}}\beta_t}{1-\bar{\alpha}_t}\mathbf{x}_0 + \frac{\sqrt{\alpha_t}(1-\bar{\alpha}_{t-1})}{1-\bar{\alpha}_t}\mathbf{x}_t.
\end{align}

\paragraph{Neural Parameterization.}
We parameterize by predicting added noise:
\begin{equation}
\boldsymbol{\mu}_\theta(\mathbf{x}_t, t) = \frac{1}{\sqrt{\alpha_t}}\left(\mathbf{x}_t - \frac{\beta_t}{\sqrt{1-\bar{\alpha}_t}}\boldsymbol{\epsilon}_\theta(\mathbf{x}_t, t)\right),
\end{equation}
where $\boldsymbol{\epsilon}_\theta$ is a U-Net predicting noise $\boldsymbol{\epsilon}$.

\paragraph{Training Objective.}
Simplified denoising score matching:
\begin{equation}
\mathcal{L}_{\text{simple}} = \mathbb{E}_{t \sim \text{Uniform}(1,T), \mathbf{x}_0, \boldsymbol{\epsilon}}\left[\|\boldsymbol{\epsilon} - \boldsymbol{\epsilon}_\theta(\mathbf{x}_t, t)\|^2\right].
\end{equation}

\paragraph{Sampling.}
Reverse diffusion from $\mathbf{x}_T \sim \mathcal{N}(\mathbf{0}, \mathbf{I})$:
\begin{equation}
\mathbf{x}_{t-1} = \frac{1}{\sqrt{\alpha_t}}\left(\mathbf{x}_t - \frac{1-\alpha_t}{\sqrt{1-\bar{\alpha}_t}}\boldsymbol{\epsilon}_\theta(\mathbf{x}_t, t)\right) + \sigma_t\mathbf{z},
\end{equation}
where $\mathbf{z} \sim \mathcal{N}(\mathbf{0}, \mathbf{I})$ for $t > 1$, else $\mathbf{z} = \mathbf{0}$.

\subsubsection{Conditional Generation}

Extend to text conditioning: $\boldsymbol{\epsilon}_\theta(\mathbf{x}_t, t, \mathbf{c})$ where $\mathbf{c} \in \mathbb{R}^{768}$ is CLIP text embedding.

\paragraph{Cross-Attention.}
At each U-Net layer with features $\mathbf{F} \in \mathbb{R}^{h \times w \times c}$ and text $\mathbf{C} \in \mathbb{R}^{L \times d}$:
\begin{align}
\mathbf{Q} &= \mathbf{W}_Q \text{Flatten}(\mathbf{F}), \quad
\mathbf{K} = \mathbf{W}_K \mathbf{C}, \quad
\mathbf{V} = \mathbf{W}_V \mathbf{C}, \\
\text{Attention} &= \text{softmax}\left(\frac{\mathbf{Q}\mathbf{K}^T}{\sqrt{d_k}}\right)\mathbf{V}.
\end{align}

\paragraph{Classifier-Free Guidance.}
Strengthen conditioning at inference:
\begin{equation}
\tilde{\boldsymbol{\epsilon}}_\theta = \boldsymbol{\epsilon}_\theta(\mathbf{x}_t, t, \emptyset) + w \cdot (\boldsymbol{\epsilon}_\theta(\mathbf{x}_t, t, \mathbf{c}) - \boldsymbol{\epsilon}_\theta(\mathbf{x}_t, t, \emptyset)),
\end{equation}
where $w > 1$ is guidance scale and $\emptyset$ denotes null conditioning.

\subsection{Latent Diffusion Models}

Operate in compressed VAE space. Pre-trained encoder $\mathcal{E}$ and decoder $\mathcal{D}$ with downsampling $f$:
\begin{equation}
\mathbf{z} = \mathcal{E}(\mathbf{x}) \in \mathbb{R}^{H/f \times W/f \times c_z}, \quad \hat{\mathbf{x}} = \mathcal{D}(\mathbf{z}) \in \mathbb{R}^{H \times W \times 3}.
\end{equation}

For Stable Diffusion: $f=8$, $c_z=4$. This provides 64× speedup per attention layer and 16-32× overall training acceleration. Latent objective:
\begin{equation}
\mathcal{L}_{\text{latent}} = \mathbb{E}_{t, \mathbf{z}_0, \boldsymbol{\epsilon}}\left[\|\boldsymbol{\epsilon} - \boldsymbol{\epsilon}_\theta(\mathbf{z}_t, t, \mathbf{c})\|^2\right].
\end{equation}

\subsection{U-Net Architecture}

Following Stable Diffusion v1.5 with random initialization (860M parameters).

\paragraph{Structure.}
Four stages at resolutions $\{h, h/2, h/4, h/8\}$ with channels $\{320, 640, 1280, 1280\}$. Each stage:
\begin{itemize}
\item 2-3 ResNet blocks with timestep injection
\item Self-attention (heads=8) for coarser resolutions
\item Cross-attention (heads=8) to CLIP text embeddings
\item Down/upsampling between stages
\end{itemize}

\paragraph{ResNet Block.}
\begin{align}
\mathbf{h} &= \text{Conv}_{3\times3}(\text{SiLU}(\text{GroupNorm}_{32}(\mathbf{F})), C_{\text{out}}), \\
\mathbf{h} &= \mathbf{h} + \text{Linear}(t_{\text{emb}}), \\
\mathbf{h} &= \text{Conv}_{3\times3}(\text{SiLU}(\text{GroupNorm}_{32}(\mathbf{h})), C_{\text{out}}), \\
\mathbf{F}_{\text{out}} &= \mathbf{h} + \text{Residual}(\mathbf{F}).
\end{align}

\paragraph{Timestep Embedding.}
Sinusoidal encoding with MLP projection to 1280-dim:
\begin{equation}
\text{PE}_i(t) = \begin{cases}
\sin(t / 10000^{2i/256}) & \text{if } i \text{ even} \\
\cos(t / 10000^{2(i-1)/256}) & \text{if } i \text{ odd}
\end{cases}.
\end{equation}

Projected via Linear(256→1024)

→SiLU→Linear(1024→1280).
\section{Implementation Details}
\label{sec:implementation}

This section provides comprehensive technical specifications for reproducibility. Our training pipeline consists of two stages with distinct configurations (as shown in Tables~\ref{tab:stage1_config} and \ref{tab:stage2_config}), executed on high-performance computing infrastructure (as shown in Table~\ref{tab:compute_full}). The baseline models for comparison are configured following their original implementations with adaptations for our evaluation protocol (see Section~\ref{subsec:baseline_models}).
\subsection{Dataset Details}
\label{subsec:dataset_details}

\paragraph{Data Collection Timeline.}
The 18-month collection phase (January 2023 - June 2024):

\noindent\textbf{Stage 1 (Months 1-6):} Integration of 20 public benchmarks (MVTec AD, VisA, BTAD, etc.)

\noindent\textbf{Stage 2 (Months 7-12):} Web mining across GitHub, RoboFlow, PaddlePaddle, Tianchi using multilingual queries in English, Chinese, Japanese. Keywords included defect detection, quality inspection. Yielded ~180K samples.

\noindent\textbf{Stage 3 (Months 13-18):} Industrial partnerships with 12 companies across petrochemical (4 companies), metal processing (5 companies), and powder metallurgy (3 companies). All data anonymized: EXIF removal, serial number blurring, facility layout suppression.

\paragraph{Annotation Protocol.}
23 expert annotators (5-15 years QC experience) following three-stage verification:

\noindent\textbf{Stage 1 - Initial Annotation:} Primary annotator creates masks and textual descriptions following template:
\begin{quote}
\small
\texttt{[Product Category] [Material] with [Defect Type] located at [Spatial Location], characterized by [Morphological Descriptors], potentially caused by [Root Cause].}
\end{quote}

\noindent\textbf{Stage 2 - Peer Review:} Secondary annotator verifies technical accuracy. Disagreements (18.3\%) flagged.

\noindent\textbf{Stage 3 - Consensus:} Panel of 3+ experts resolves conflicts. Highly ambiguous cases (2.7\%) undergo additional inspection.

Inter-annotator agreement: Cohen's $\kappa = 0.87$ (classification), $\kappa = 0.81$ (segmentation IoU>0.75).

\paragraph{Representative Qualitative Examples.}

To contextualize our dataset and illustrate the breadth of visual patterns it captures, we present representative defect samples from multiple industrial domains (Figures~\ref{fig:alu} and \ref{fig:llm}). The first group focuses on aluminum surfaces, exhibiting subtle yet distinct failure modes that vary widely in morphology and optical response. The broader cross-material selection spans photovoltaics, metallic alloys, packaging components, precision gears, and textile fibers, including microcracks, oxidation, pin holes, dents, and fiber breakage, demonstrating the dataset’s diversity and realism across industrial settings.

\subsection{Compared Model Settings}
\label{subsec:baseline_models}

\paragraph{YOLOv8-m Configuration.}
For object detection comparison (Table 4, main paper):
\begin{itemize}
\item Architecture: CSPDarknet53 backbone, PANet neck, decoupled head
\item Parameters: 25.9M (backbone: 13.2M, neck: 8.4M, head: 4.3M)
\item Input: $640 \times 640$ pixels with letterbox padding
\item Training: 300 epochs, batch size 16, early stopping (patience=50)
\item Optimizer: SGD (momentum 0.937, weight decay $5 \times 10^{-4}$)
\item Learning rate: $10^{-2}$ initial, cosine decay to $10^{-5}$, 3 epochs warmup
\item Loss: CIoU (7.5) + DFL (1.5) + BCE (0.5)
\item Augmentation: Mosaic (0.8), Mixup (0.15), HSV jitter, flip (0.5)
\item Training time: 8 hours on 4× RTX 3090 (32 GPU-hours)
\item Inference: 6.2ms per image on A100 (161 FPS)
\end{itemize}

\clearpage
\begin{strip}
\centering
\includegraphics[width=0.95\textwidth]{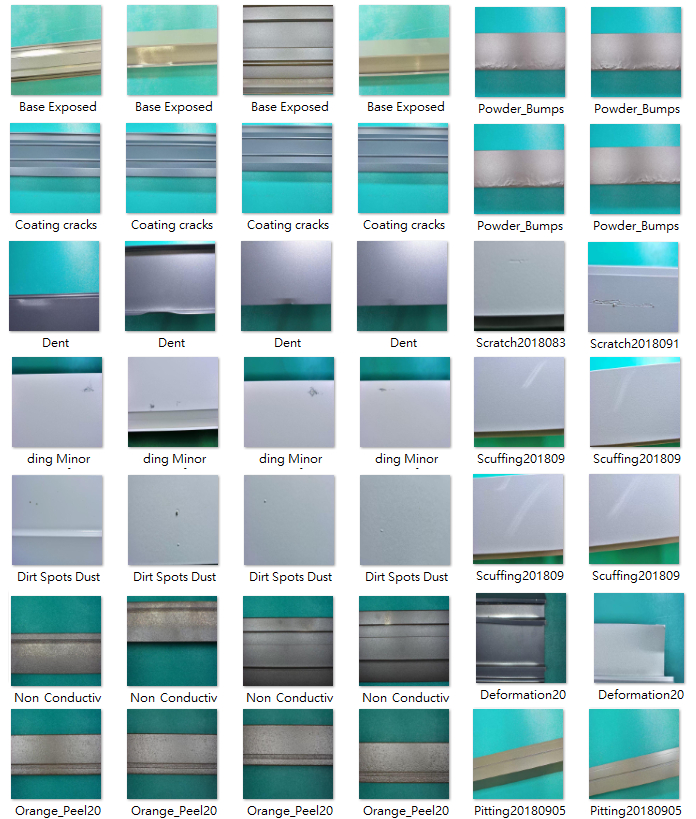}
\vspace{-0.5em}
\captionof{figure}{
This figure showcases a diverse set of aluminum surface defects, including
base-exposed regions, coating cracks, powder bumps, dents, scratches,
minor dings, dust spots, scuffing marks, non-conductive stripes, deformation
artifacts, orange-peel textures, and pitting defects. The samples highlight
variations in texture, reflectivity, and severity, providing a comprehensive
visual reference for real-world aluminum anomaly patterns.
}
\label{fig:alu}
\end{strip}

\clearpage
\begin{strip}
\centering
\includegraphics[width=0.75\textwidth]{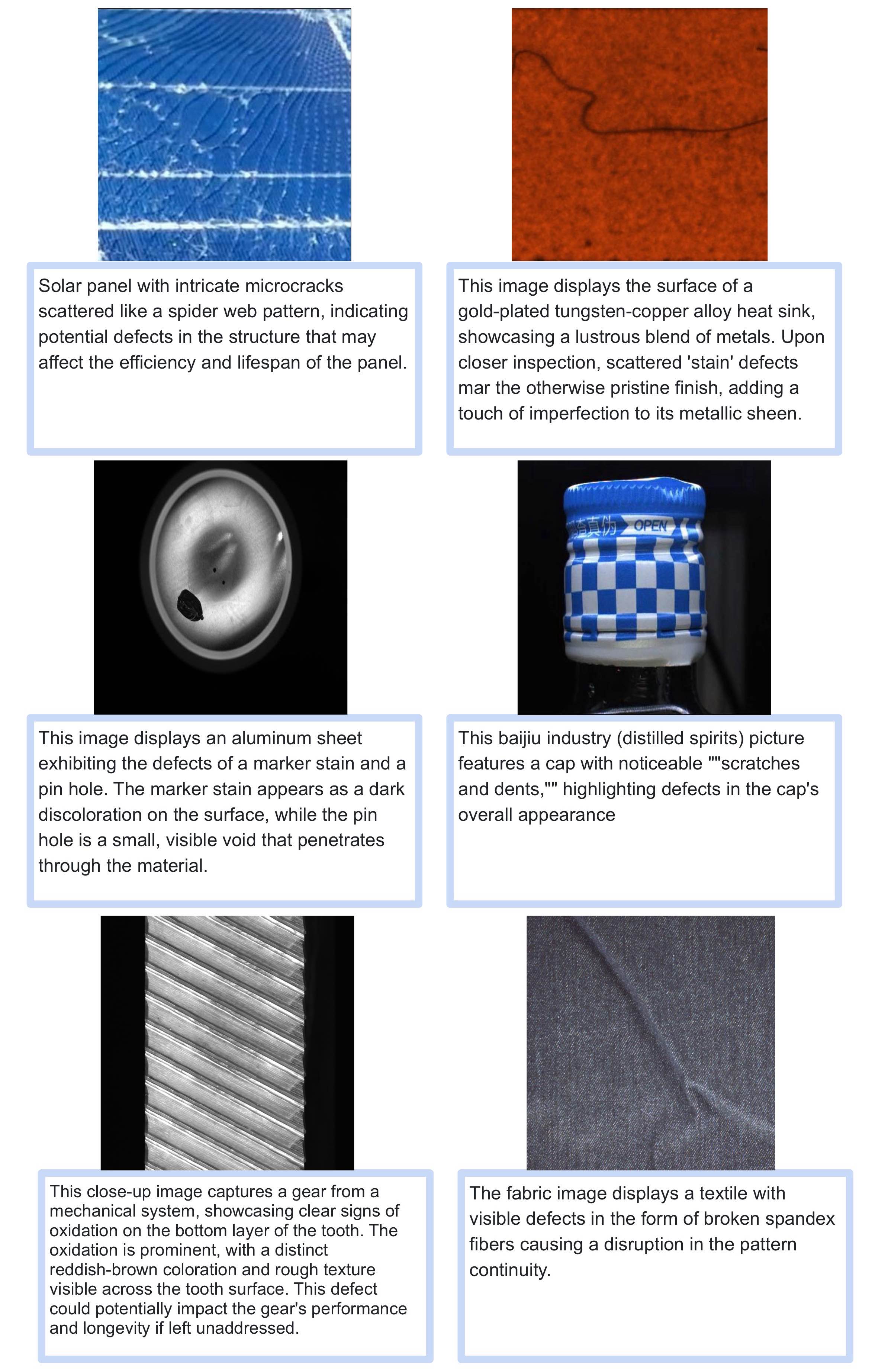}
\vspace{-0.5em}
\captionof{figure}{
A diverse collection of real-world defect examples across multiple material domains, including microcracks in solar panels, surface streaks on metallic alloys, stains and pin holes on aluminum sheets, scratches and dents on bottle caps, oxidation on mechanical gears, and fiber breakage in textile fabrics. These samples highlight the wide variability in appearance, texture, and failure modes encountered in practical industrial settings.
}
\label{fig:llm}
\end{strip}

The complete Stage 1 pre-training hyperparameters are detailed in Table~\ref{tab:stage1_config}, while Stage 2 fine-tuning settings are provided in Table~\ref{tab:stage2_config}.

\paragraph{Anomaly Detection Baselines.}
For segmentation comparison (Table 6, main paper):

\noindent\textbf{1. MuSc (Mutual Scoring):}
ViT-B/16 pre-trained on ImageNet-21K. Self-supervised contrastive learning on unlabeled normals. 86M frozen backbone + 2.3M trainable projection. 100 epochs, batch size 32, lr=$10^{-3}$.

\noindent\textbf{2. PromptAD:}
CLIP ViT-B/16 with learnable prompts. 86M frozen encoders + 0.8M prompts (10 tokens per category). 4-shot (4 normal samples), 50 epochs, AdamW lr=$10^{-4}$.

\noindent\textbf{3. DMAD (Diversity-Measurable):}
Diffusion U-Net 250M params. 200 epochs per category (3000 total for MVTec AD). 100 diffusion steps, cosine schedule. Unsupervised (normal samples only). ~15 hours per category on 8× A100.

\noindent\textbf{4. SimpleNet:}
WideResNet-50 frozen + 12M adaptation network. 150 epochs, batch size 32, lr=$10^{-3}$. Mahalanobis distance to memory bank. 2.5 hours per category on 4× RTX 3090.

\noindent\textbf{5. FAIR (Frequency-Aware):}
Dual-branch (spatial 25M + frequency 18M + fusion 2M). 200 epochs, batch size 16, lr=$5 \times 10^{-4}$. Loss: L1 (1.0) + perceptual (0.1) + frequency (0.5). 6 hours per category on 4× A100.

All baselines trained on full MVTec AD (~4000 samples per class after 20× augmentation).

\subsection{Training Details}
\label{subsec:training_details}

\paragraph{Stage 1: Diffusion U-Net Pre-training.}

\begin{table}[h]
\centering
\small
\begin{tabular}{lc}
\toprule
\textbf{Configuration} & \textbf{Value} \\
\midrule
Optimizer & AdamW \\
Base Learning Rate & $1 \times 10^{-4}$ \\
Weight Decay & $1 \times 10^{-4}$ \\
Optimizer Momentum & $\beta_1=0.9$, $\beta_2=0.999$ \\
Batch Size & 256 (32 per GPU × 8) \\
Learning Rate Schedule & Cosine Decay \\
Warmup Steps & 5,000 iterations \\
Training Epochs & 100 \\
Gradient Clipping & Max norm 1.0 \\
EMA Decay & 0.9999 \\
Mixed Precision & FP16 (AMP) \\
\midrule
Diffusion Steps $T$ & 1000 \\
Noise Schedule & Linear \\
$\beta_1$ (start) & $1 \times 10^{-4}$ \\
$\beta_T$ (end) & $2 \times 10^{-2}$ \\
\midrule
Loss Weight $\mathcal{L}_{\text{diff}}$ & 1.0 \\
Loss Weight $\mathcal{L}_{\text{imp}}$ & 0.3 \\
Text Conditioning Probability & 0.5 / 0.5 \\
\midrule
\textbf{Augmentation} & \\
Random Horizontal Flip & $p = 0.5$ \\
Random Vertical Flip & $p = 0.5$ \\
Random Rotation & $\pm 15°$ \\
Color Jitter (BSCH) & 0.2, 0.2, 0.1, 0.05 \\
Random Resized Crop & scale=[0.8, 1.0] \\
\bottomrule
\end{tabular}
\caption{Stage 1 diffusion U-Net pre-training configuration on IMDD-1M.}
\label{tab:stage1_config}
\end{table}

\noindent\textbf{Training Details:} U-Net trained from random initialization (He for conv, Xavier for linear). Warmup: 5000 steps from $10^{-6}$ to $10^{-4}$. Cosine decay to $10^{-6}$. Implicit captioner trained jointly with stochastic conditioning. EMA updated every iteration.

\noindent\textbf{Resources:} 72 hours on 8× H100 80GB, 576 GPU-hours total. Peak memory 76GB/GPU. ~43 min/epoch, 484,500 total iterations.

\paragraph{Stage 2: Mask Generator Fine-tuning.}

\begin{table}[h]
\centering
\small
\begin{tabular}{lc}
\toprule
\textbf{Configuration} & \textbf{Value} \\
\midrule
Optimizer & AdamW \\
Base Learning Rate & $5 \times 10^{-5}$ \\
Weight Decay & $1 \times 10^{-4}$ \\
Batch Size & 16 (2 per GPU × 8) \\
LR Schedule & Polynomial Decay (power=0.9) \\
Warmup Steps & 500 iterations \\
Training Epochs & 50 \\
Gradient Clipping & Max norm 0.01 \\
Mixed Precision & FP16 \\
\midrule
Feature Timestep $t$ & 50 \\
Loss Weight $\mathcal{L}_{\text{mask}}$ & 1.0 \\
Loss Weight $\mathcal{L}_{\text{cls/ground}}$ & 0.5 \\
Mask Queries & 100 \\
Transformer Layers & 9 \\
\midrule
\textbf{Frozen Components} & \\
Diffusion U-Net & 860M params \\
VAE Encoder/Decoder & 84M params \\
CLIP & 63M params \\
Implicit Captioner & 0.3M params \\
\midrule
\textbf{Trainable Components} & \\
Mask2Former & 45M params \\
\bottomrule
\end{tabular}
\caption{Stage 2 mask generator fine-tuning configuration.}
\label{tab:stage2_config}
\end{table}

\noindent\textbf{Training Time:} MVTec AD (3629 samples): 4 hours. VisA (9621 samples): 5.5 hours.

\subsection{Computing Resource Configuration}
\label{subsec:compute_resources}

\begin{table}[h]
\centering
\small
\begin{tabular}{ll}
\toprule
\textbf{Component} & \textbf{Specification} \\
\midrule
\multicolumn{2}{l}{\textit{\textbf{Hardware}}} \\
GPU & 8× NVIDIA H100 80GB \\
CPU & 2× AMD EPYC 7763 (128 cores) \\
RAM & 2TB DDR4-3200 ECC \\
Storage & 100TB NVMe SSD RAID-0 \\
\midrule
\multicolumn{2}{l}{\textit{\textbf{Software}}} \\
OS & Ubuntu 22.04 LTS \\
CUDA & 12.1 \\
PyTorch & 2.1.0 \\
Python & 3.10 \\
\midrule
\multicolumn{2}{l}{\textit{\textbf{Inference}}} \\
Latency & 0.35s per image (A100) \\
Throughput & 2.86 images/sec \\
Memory & 18.7 GB \\
\bottomrule
\end{tabular}
\caption{Computing resource configuration.}
\label{tab:compute_full}
\end{table}

\paragraph{Memory Optimization:}
Gradient checkpointing (40\% memory saving), mixed precision FP16, gradient accumulation for smaller GPUs.

\paragraph{Distributed Training:}
PyTorch DDP with NCCL backend. 32 data loader workers per GPU (256 total). NVLink 4.0 for gradient all-reduce (~150ms, overlapped down to 40ms).

\section{Additional Experiments}
\label{sec:additional_experiments}

\subsection{Extended Quantitative Analysis}

\paragraph{Per-Category Performance.}
We evaluate DiffuseDefect across 10 MVTec AD categories (200 samples/class). As shown in Table~\ref{tab:per_category_full}, our method achieves a remarkable $\mathbf{91.99\%}$ average accuracy (Acc) and $\mathbf{55.71\%}$ mean IoU. Results are consistent across types, including Grid ($\mathbf{94.32\%}$ Acc, $\mathbf{61.2\%}$ IoU), Leather ($\mathbf{93.67\%}$ Acc, $\mathbf{59.7\%}$ IoU), and Cable ($\mathbf{89.70\%}$ Acc, $\mathbf{51.4\%}$ IoU).

\begin{table}[h]
\centering
\small
\begin{tabular}{lccc}
\toprule
\textbf{Category} & \textbf{Acc (\%)} & \textbf{F1 (\%)} & \textbf{IoU (\%)} \\
\midrule
Grid & 94.32 & 67.8 & 61.2 \\
Leather & 93.67 & 65.2 & 59.7 \\
Cable & 89.70 & 56.8 & 51.4 \\
\midrule
\textbf{Average} & \textbf{91.99} & \textbf{61.48} & \textbf{55.71} \\
\bottomrule
\end{tabular}
\caption{Per-category results on MVTec AD (200 samples/class).}
\label{tab:per_category_full}
\end{table}

\paragraph{Cross-Dataset Generalization.}
As shown in Table~\ref{tab:cross_dataset}, IMDD-1M pre-trained models achieve zero-shot transfer IoU of $\mathbf{52.9\%}$ to $\mathbf{54.7\%}$, providing an $\mathbf{11\%}$ to $\mathbf{15\%}$ IoU gain over single-dataset baselines.

\begin{table}[h]
\centering
\small
\begin{tabular}{lcc}
\toprule
\textbf{Train $\rightarrow$ Test} & \textbf{Acc (\%)} & \textbf{IoU (\%)} \\
\midrule
MVTec AD $\rightarrow$ VisA & 83.2 & 41.3 \\
IMDD-1M $\rightarrow$ MVTec AD & \textbf{91.0} & \textbf{52.9} \\
IMDD-1M $\rightarrow$ VisA & \textbf{90.3} & \textbf{54.7} \\
\bottomrule
\end{tabular}
\caption{Zero-shot cross-dataset transfer performance.}
\label{tab:cross_dataset}
\end{table}

\subsection{Ablation Studies}

\paragraph{Training from Scratch vs. Fine-tuning.}
We compare training from random initialization versus fine-tuning from pre-trained Stable Diffusion weights. Random initialization achieves $\mathbf{82.7\%}$ mIoU, outperforming fine-tuned Stable Diffusion ($\mathbf{74.5\%}$) by $\mathbf{8.2\%}$. This indicates that natural image priors may actually hinder learning of industrial defect patterns, which have fundamentally different visual characteristics.

\paragraph{Timestep Selection.}
We investigate the impact of the diffusion timestep $t$ on feature extraction quality. Our analysis shows that timestep $t=50$ provides the optimal balance between semantic understanding and spatial precision, achieving $\mathbf{91.0\%}$ accuracy and $\mathbf{52.9\%}$ IoU. Earlier timesteps preserve more spatial detail but lack semantic context, while later timesteps capture high-level semantics but lose fine-grained localization.

\paragraph{Sample Efficiency.}
We evaluate the data efficiency of IMDD-1M pre-training by measuring the samples required to reach $\mathbf{95\%}$ accuracy. As shown in Table~\ref{tab:sample_efficiency}, IMDD-1M pre-training requires only $\mathbf{150}$ samples, which is $\mathbf{12.3\times}$ more efficient than random initialization ($\mathbf{1850}$ samples) and $\mathbf{3.6\times}$ more efficient than ImageNet ($\mathbf{520}$ samples). This improvement highlights the value of domain-specific pre-training.

\begin{table}[h]
\centering
\small
\begin{tabular}{lcc}
\toprule
\textbf{Pre-training} & \textbf{Samples for 95\% Acc} & \textbf{Efficiency} \\
\midrule
Random Init & 1850 & 1.0$\times$ \\
ImageNet & 520 & 3.6$\times$ \\
\textbf{IMDD-1M} & \textbf{150} & \textbf{12.3$\times$} \\
\bottomrule
\end{tabular}
\caption{Sample efficiency comparison across different pre-training strategies.}
\label{tab:sample_efficiency}
\end{table}

\subsection{Dataset Statistics}
Annotation achieved Cohen's $\kappa$ of $\mathbf{0.87}$ (classification) and $\mathbf{0.81}$ (segmentation), requiring $\mathbf{66,287}$ hours. The defect distribution exhibits a realistic long-tail: top 10 types comprise $\mathbf{47.4\%}$, while $\mathbf{411}$ rare types account for $\mathbf{52.6\%}$.

\subsection{Real and Generated Visual Comparison}
To further illustrate the visual diversity in our dataset and evaluate the generative model's effectiveness, we present qualitative comparisons between real and synthesized defect images (Figure~\ref{fig:air}).

\clearpage
\begin{strip}
\centering
\includegraphics[width=0.95\textwidth]{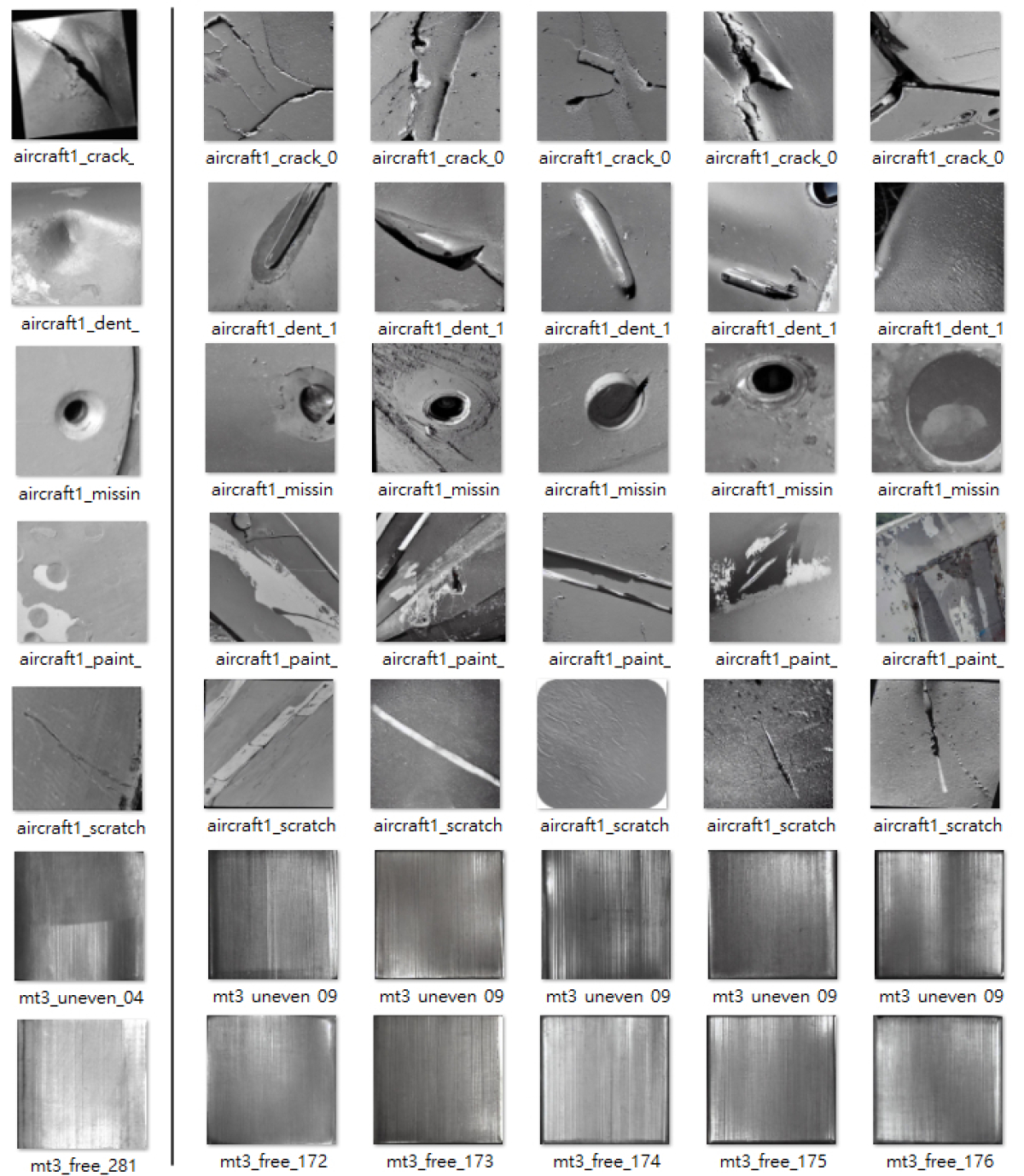}
\vspace{-0.5em}
\captionof{figure}{
Comparison between real defect samples and model-generated counterparts across various aircraft and metal-surface categories, including cracks, dents, missing regions, paint defects, scratches, and uneven textures. The generated images closely reproduce the structural morphology, surface patterns, and material appearance observed in the real samples, illustrating the model’s ability to synthesize realistic defect characteristics. Each column pair shows a real sample on the left and the corresponding generated sample on the right.
}
\label{fig:air}
\end{strip}


\end{document}